\documentclass[usletter, 10 pt, conference]{ieeeconf}  

\usepackage{graphicx}
\usepackage{amsmath,amssymb} 
\usepackage{color}
\usepackage{algorithm}
\usepackage{algpseudocode}
\usepackage{enumerate}
\usepackage{epstopdf}
\usepackage{array}
\usepackage{url}
\usepackage{gensymb}
\IEEEoverridecommandlockouts                              

\newcommand{\qed}{\nobreak \ifvmode \relax \else
      \ifdim\lastskip<1.5em \hskip-\lastskip
      \hskip1.5em plus0em minus0.5em \fi \nobreak
      \vrule height0.75em width0.5em depth0.25em\fi}

\title{\LARGE \bf
Exploring Convolutional Networks for End-to-End Visual Servoing
}

\author{Aseem Saxena*$^1$, Harit Pandya*$^1$, Gourav Kumar$^1$, Ayush Gaud$^1$ and K. Madhava Krishna$^1$ 
\thanks{*Equal contribution.}
\thanks{$^1$ International Institute of Information Technology, Hyderabad, India.}
\thanks{\{harit.pandya, gourav.kumar\}@research.iiit.ac.in}
\thanks{\{aseem.bits, ayush.gaud\}@gmail.com, mkrishna@iiit.ac.in}
\thanks{Harit Pandya is supported by TCS Reserach PhD fellowship.}
}

\begin{document}

\maketitle
\thispagestyle{empty}
\pagestyle{empty}

\begin{abstract}
 Present image based visual servoing approaches rely on extracting hand crafted visual features from an image. Choosing the right set of features is important as it directly affects the performance of any approach. Motivated by recent breakthroughs in performance of data driven methods on recognition and localization tasks, we aim to learn visual feature representations suitable for servoing tasks in unstructured and unknown environments. In this paper, we present an end-to-end learning based approach for visual servoing in diverse scenes where the knowledge of camera parameters and scene geometry is not available a priori. This is achieved by training a convolutional neural network over color images with synchronised camera poses. Through experiments performed in simulation and on a quadrotor, we demonstrate the efficacy and robustness of our approach for a wide range of camera poses in both indoor as well as outdoor environments. 
\end{abstract}
\section{INTRODUCTION}
\label{sec:intro}
Visual servoing (VS) refers to the control of robot motion using data from vision sensors. Vision sensor integration enables robotic systems to work outside controlled industrial settings. As a consequence, it has applications in diverse areas such as robotic surgery, autonomous navigation and manipulation for household robots. The objective is to move the robot in Cartesian space from an arbitrary starting pose (location and orientation) to a fixed goal pose. This is achieved by iteratively minimizing the error between the current and goal pose.\\ 
\indent Position Based Visual Servoing (PBVS) defines the error in Euclidean space, which results in a simpler control law and minimal length trajectory \cite{vsbasic}. However, PBVS requires a $3$D model of the scene and camera parameters to be known before hand, which is a major bottleneck in practical implementation of PBVS methods. Image based visual servoing (IBVS), on the other hand, 
describes the error function in image space by extracting a set of visual features. IBVS controller attempts to move the robot in such a way that the visual features attain the desired configuration i.e. it minimises the error in image space.  This requires a mapping of the feature velocity in image space to robot motion in Euclidean space via the analytical computation of image Jacobian that leads to various issues such as attaining local minima, exceeding joint limit and so forth \cite{vsprob}. Another issue with IBVS approaches is the extraction of unambiguous features that truly represent the pose information, which  is a non-trivial task. Classical IBVS approaches use geometrical primitives like locations of points, lines, contours etc. as visual features \cite{vsbasic}. However, these methods require an accurate feature matching for convergence. Recent IBVS methods consider appearance based features such as pixel intensities \cite{photometricvs}, image gradients \cite{gradientvs} etc. These methods do not require an explicit matching step, however, the number of features is typically very large that results in a smaller convergence domain.\\
\begin{figure}[t!]
\begin{center}
\begin{tabular}{c}
     \framebox{\includegraphics[width=7.0cm, height=3.1cm]{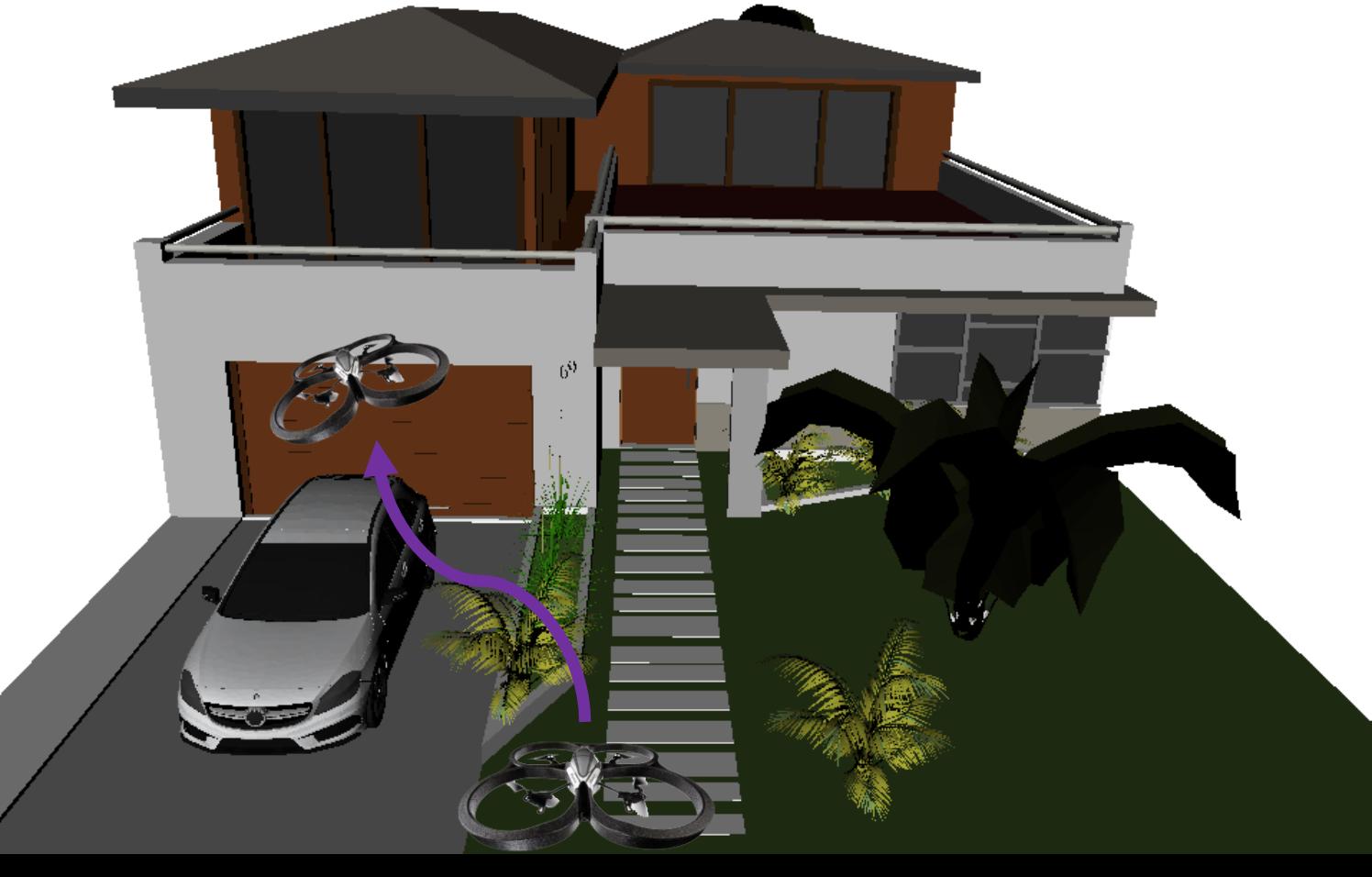}} \\
     (a)\\
     \begin{tabular}{m{2.33cm}m{2.33cm}m{2.33cm}}
     \framebox{\includegraphics[width=2.33cm, height=2.8cm]{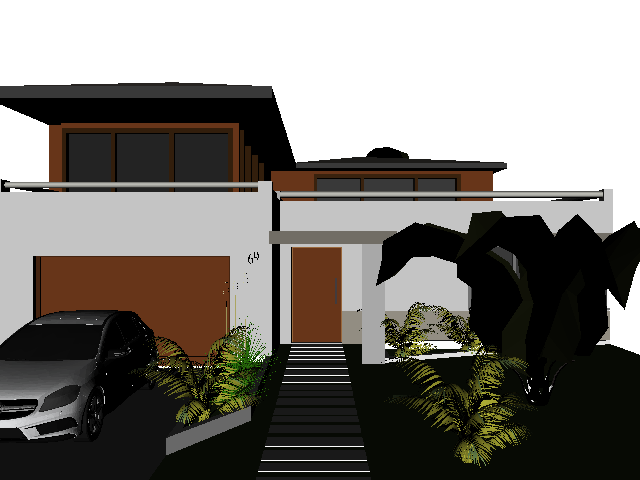}} &
     \framebox{\includegraphics[width=2.33cm, height=2.8cm]{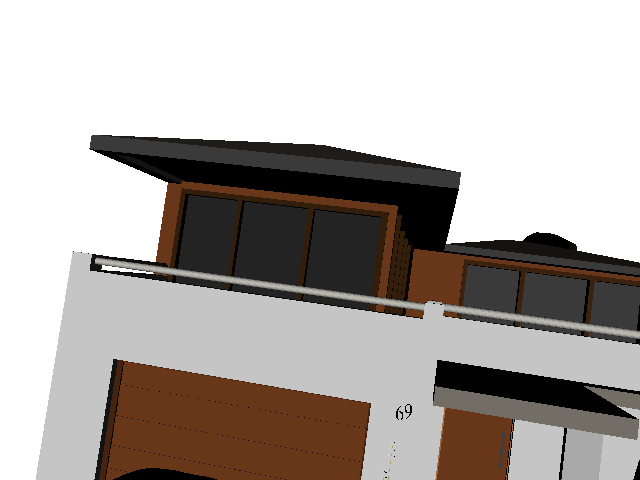}} &
     \framebox{\includegraphics[width=2.33cm, height=2.8cm]{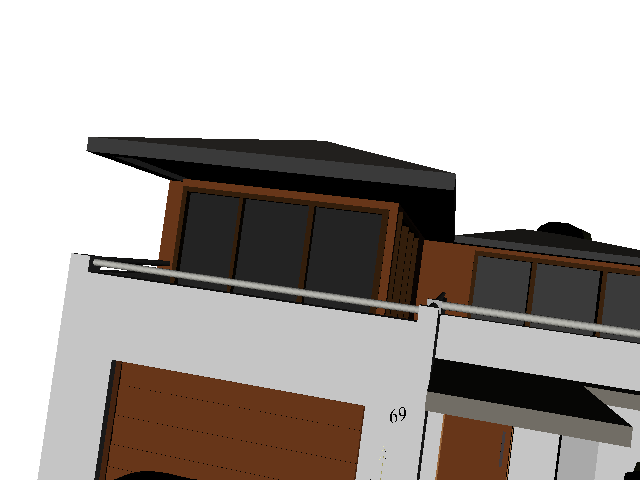}} \\
     \centering (b) & \centering (c) & \centering (d)
     \end{tabular}
\end{tabular}

\caption{\textbf{Visual servoing using CNN.} We present a learning based approach for end-to-end visual servoing using CNNs i.e. given a desired pose and initial pose in the form of images, our framework predicts visual servoing control commands. (a) A $3$D model of an outdoor scene from our dataset with an initial and desired location of the robot.  (b) Initial pose. (c)  Desired pose. (d) Resultant pose attained by the robot using our framework. Note that our network does not assume prior knowledge of camera parameters and geometry of the scene. Also note the large camera displacement between initial and desired pose.}
\label{fig:probstat}
\end{center}
\vspace{-1.75em}
\end{figure}
\indent 
In this paper, we address the following question - is it possible to learn the motion required to attain a desired pose from an initial pose only from visual feedback? Recent breakthroughs in computer vision suggest that data driven frameworks efficiently learn high-level semantic representations from images, especially for a large number of examples \cite{krizhevsky2012imagenet}. Motivated by recent advances in machine learning, especially deep learning, we aim to learn an optimal set of image representations that estimate the relative transformation required to attain a desired pose. We explore convolutional neural network (CNN) architectures to learn such transformations in an end-to-end paradigm. Unlike other visual servoing approaches, our framework eliminates the need for extraction and tracking of features. Moreover, prior knowledge about the camera intrinsics and the scene's $3$D geometry is not required. Experiments show that our model also has a large convergence domain across a variety of synthetic and real world scenarios.\\
\indent In this paper, we present a Convolutional Neural Network trained for performing visual servoing on diverse environments without knowledge of the underlying scene's geometry. We have trained the network on the publicly available 7-Scenes dataset \cite{glocker2013real} as this dataset provides large variations across scenes and covers a  wide range of camera transformations between frames. We evaluate our network on $5$ synthetic $3$D models using free camera paradigm and on a real world scene using a quadrotor. Our simulation based testing framework allows us to compute ground truth camera transformations that can be used to compute the error of our system's estimates. Figure \ref{fig:probstat} shows an exemplar of servoing result.  Figure \ref{fig:probstat}(a) shows the scene on which servoing was performed. Figure \ref{fig:probstat}(b-d) shows initial pose, desired pose and resultant pose attained after visual servoing. Note that although there is large camera motion between initial and desired pose, the camera still reaches close to the desired pose using our method. 

\section{Related work}
Most of the previous image based visual servoing approaches rely on hand-crafted visual features for representing images. The control law could be seen as gradient descent over the feature error \cite{photometricvs}. This requires image Jacobian or interaction matrix to be  computed analytically. For several features widely used by modern computer vision techniques, it is difficult to represent analytically, for instance, Histogram of Oriented Gradients (HOG). Another line of approaches intend to numerically estimate the interaction matrix. However, due to high non-linearity in interaction matrix, it is difficult to get an accurate estimation. Also, numerical methods are vulnerable to conditioning and singularity issues. Neural Network based methods have been used for learning interaction matrix but the selection of features was hand-engineered. Readers may refer to \cite{vsadvanced} for a detailed review of Jacobian learning and estimation methods.  Recently, support vector machines have been used to learn pose specific representations for visual servoing across object instances \cite{harit}. Again, the interaction matrix was numerically estimated. There has been significant work on reinforcement learning (RL) approaches \cite{RL1}, \cite{RL2}  for end-to-end visual servoing. However, parameters learned by RL are specific to the environment and task, hence it becomes difficult to generalise RL for new environments. On the other hand, our approach is end-to-end i.e. we learn visuomotor representations for direct control. Moreover, our approach generalises well on unknown environments.\\
\indent Techniques for pose estimation, camera relocalization and visual odometry have been successfully applied in approaching the visual servoing problem.
There have been works on absolute scene 6D pose estimation from a single monocular image in the recent past which are data-driven \cite{brachmannuncertainty}, \cite{kendall2015posenet}. Kendal' et al. \cite{kendall2015posenet} train a CNN
to regress the $6$D pose of the camera from a single monocular image in real time. 
Our approach differs from theirs as we wish to learn relative camera pose from a pair of images. As natural scenes change over time, systems which estimate the absolute pose of a scene are bound to falter as a viewpoint can be remarkably different visually from the same viewpoint at a different time. Rather, we consider the relative pose between two frames to be much more meaningful. Two images with sufficient scene overlap offer more information and context than a single image. Some recent works have approached the problem of camera ego-motion estimation which has applications in visual servoing \cite{agrawal2015learning}, \cite{costante2016exploring}, \cite{konda2015learning}. Agarawal et al. \cite{agrawal2015learning} explore the idea of feature learning using egomotion as ground truth instead of manually annotated labels. They demonstrate camera ego-motion estimation by learning a Siamese Style CNN with two images as input and the relative camera transformation as the ground truth. However, our work significantly differs from theirs in multiple ways. We perform regression over the image pairs whereas they perform classification. Also, we use a different Network architecture, loss function and optimization scheme for our task. Costante et al. \cite{costante2016exploring} train a network to estimate frame to frame visual odometry by taking the optical flow between image pairs as input. Ours does not require the computation of optical flow. To the best of our knowledge, there has been no work which directly addresses the problem of visual servoing by leveraging powerful CNN based image features.\\  
\indent Contributions: Our contributions could be summarised as following.
 Firstly, we present a CNN based learning framework for visual servoing. Our framework generalises well over a wide range of synthetic and real world scenarios. We rigorously and systematically evaluate our approach in simulation and on real scenarios using a quadrotor.
 Secondly, as there are no benchmarking datasets for visual servoing and due to the presence of dynamic control, it is not feasible to provide access to pre-capture images similar to most of the available datasets. We would publicly release $3$D models of scenes used for testing along with the necessary scripts.

\begin{figure}[t!]
\begin{center}
\includegraphics[width=8.6cm, height=4.8cm] {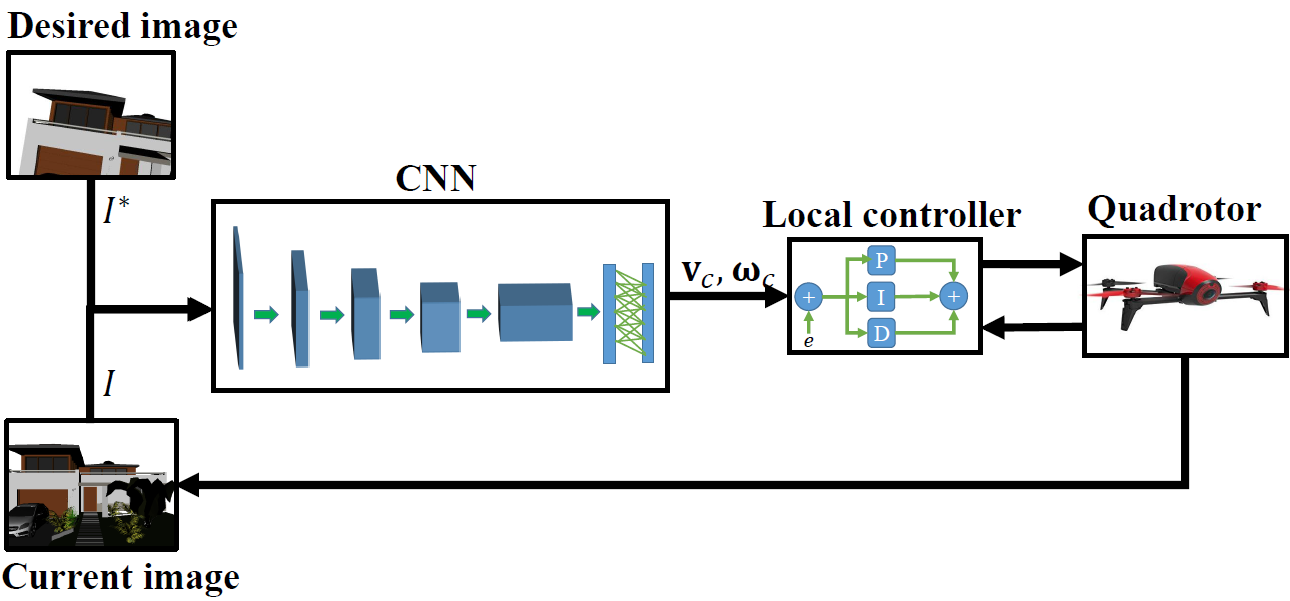}
\vspace{-1.50em}
\caption{\textbf{Overview of the proposed approach.} Given an image $I^*$ representing desired pose in the image space and the current image measurement $I$ from camera, we use a CNN to estimate relative camera transformation $_{c^*}^{c}\textrm{\textbf{T}}$ required to reach $I^*$ in image space. Considering noise in estimation of $_{c^*}^{c}\textrm{\textbf{T}}$, we take a servoing step of length $\lambda$ in direction of $_{c^*}^{c}\textrm{\textbf{T}}$. }
\label{fig:overview}
\end{center}
\vspace{-1.0em}
\end{figure}
\section{Overview}
Assuming eye-in-hand configuration and world origin coinciding with the given object's center, we denote the camera's pose in SE(3) at given time in a fixed global frame as $\textbf{c}$. Given a scene $X$ and a desired camera pose in the same global frame $c^*$, the goal of a visual servoing scheme is to find a camera transformation $_{c}^{c*}\textrm{T}$, such that  $c^* = _{c}^{c^*}\textrm{T}c$. For image based visual servoing (IBVS), current pose and the desired pose are represented in the form of a set of features extracted from images $\textbf{s}=\phi(KcX)$ and $\textbf{s}^\textbf{*}=\phi(Kc^*X)$. Where $K$ is the camera's intrinsic matrix and $\phi(\cdot)$ is the feature selection criterion. For IBVS, the goal is modified to finding the transformation $_{c}^{c*}\textrm{T}$ such that the error in features $\textbf{e}=\textbf{s}-\textbf{s}^\textbf{*}$ is regulated to zero at desired pose. The task is achieved by minimizing $\textbf{e}$ iteratively and controlling the camera velocity,  $\textbf{v} = - \lambda \textbf{L}_{\textbf{s}}^+ \textbf{e}$, where $\textbf{L}_{\textbf{s}}$ is the interaction matrix that maps the rate of change of features to velocity and $(\cdot)^+$ represents pseudoinverse operation as defined in \cite{vsbasic}. On the other hand, in  position based visual servoing (PBVS), the camera pose at any given time $c$ is inferred from the scene and image measurements. However, inferring camera pose from a single image requires the knowledge of the scene and camera parameters.
\indent In this work, we aim to jointly learn the representations $\textbf{s}$ the describe the image and error $\textbf{e}$ from a pair of images $I$ and $I^*$ without camera parameters $K$ and geometry of the scene $X$. Our framework takes RGB images $I$ and $I^*$ as the input and estimates the camera transformation $_{c*}^{c}\textrm{T}$ thereby waving the requirement for any feature computing or matching step. Further, using image based feedback, we estimate the control commands $\hat{_{c*}^{c}\textrm{T}}$  to attain the desired camera pose $c^*$. Figure \ref{fig:overview} shows the overall pipeline of the proposed framework.

\section{Network Architecture}
\begin{figure*}[t!]
\begin{center}
\begin{tabular}{cc}
\includegraphics[width=8.7cm, height=2.9cm] {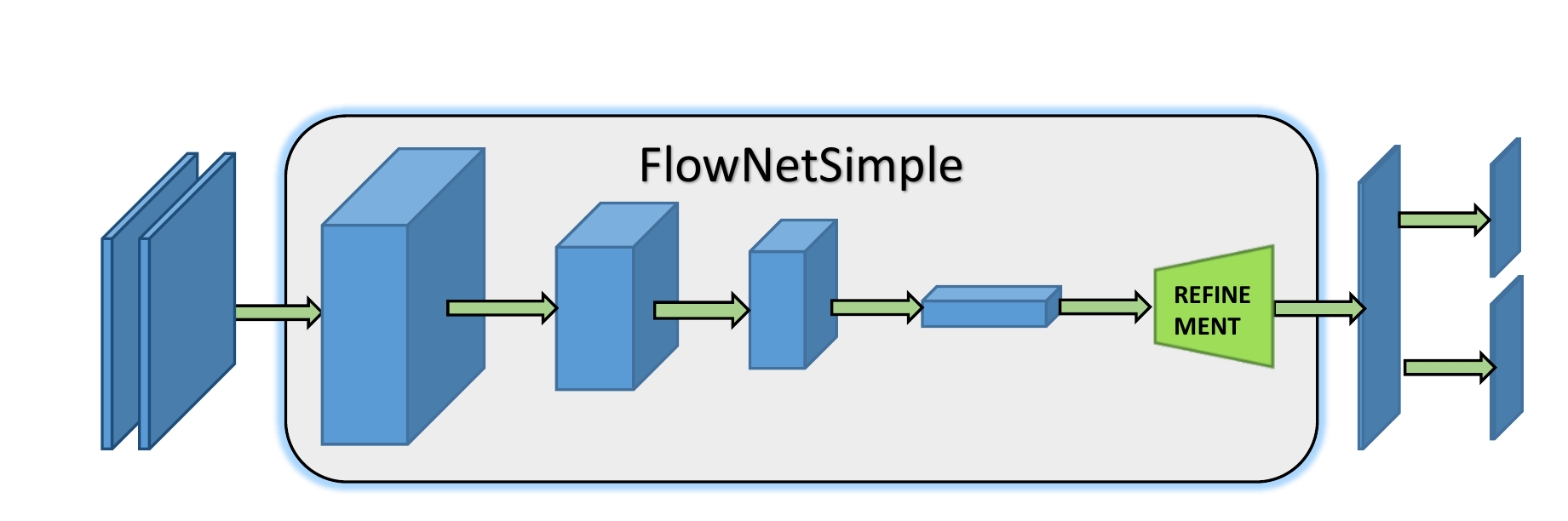} &
\includegraphics[width=8.7cm, height=2.9cm] {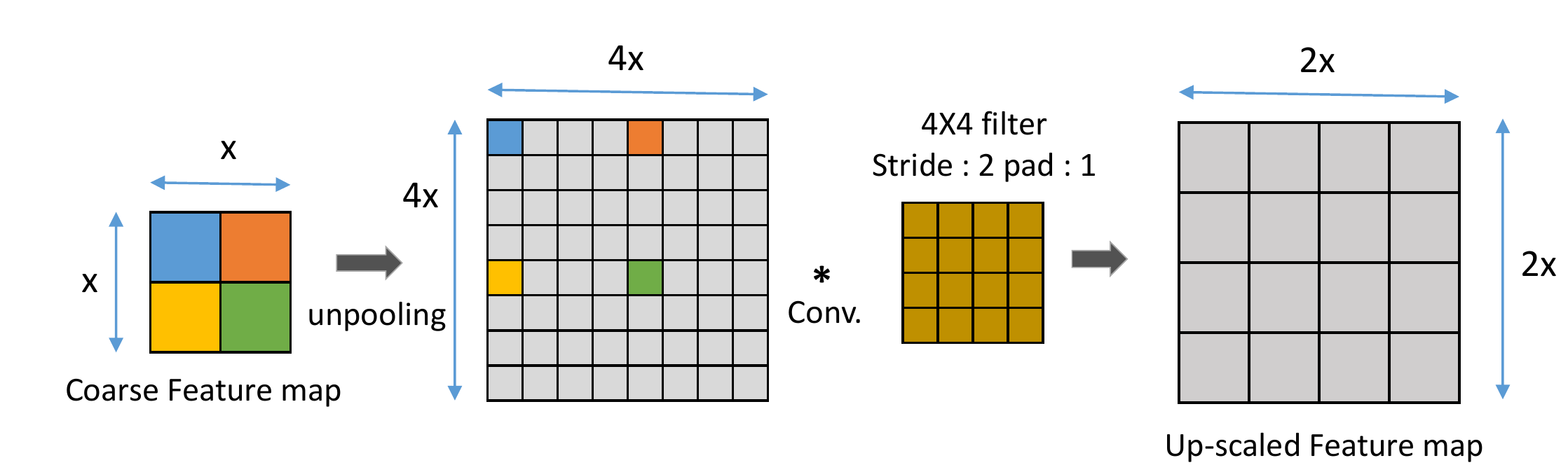} \\
(a) & (b) \\
\end{tabular}
\end{center}
\caption{\textbf{Description of the Network architectures for learning camera motion.} (a) FlowNet $-$ FlowNet takes 2 images stacked together as a six channel image as input and predicts the transformation between the two images. FlowNet consists of Convolutional, ReLu and 'Upconvolution' layers. (b) Upconvolution layer $-$ the coarse feature map is bi-linearly up-sampled to 4 times its size followed by a convolution with a $4 \times 4$ filter with stride as 2 and padding as 1. This results in a feature map double the size of the coarse feature map. Upconvolution is performed 4 times at multiple scales.  }
\label{fig:exp1}
\vspace{-0.5em}
\end{figure*}

\subsection{FlowNet}
Convolutional Neural Networks have recently been shown to perform well on large scale visual recognition tasks \cite{krizhevsky2012imagenet}. In the recent past, research on training CNNs for per pixel prediction tasks such as optical flow has started to surface \cite{dosovitskiy2015flownet}. FlowNet by Fischer et al. \cite{dosovitskiy2015flownet} approaches the problem of optical flow in a supervised learning setting. Optical flow prediction involves both per pixel localization and learning powerful representations. We leverage these aspects of FlowNet to learn camera ego-motion. The motivation behind this effort is that traditionally optical flow has been used to estimate visual odometry \cite{campbell2004techniques}. A network which could robustly estimate optical flow would also be able to estimate camera ego-motion since both problems involve correspondences between image pixels. FlowNet is trained to predict optical flow using image pairs as input and their x-y flow fields as ground truth (Figure \ref{fig:exp1}(a)). The images are stacked together to form a 6 channel image which is passed through multiple convolutions and ReLu non-linearities. Convolutional Neural Networks involve downscaling of feature maps, which is necessary for the training phase to be computationally feasible. As optical flow is a per pixel prediction task, it would require a feature map which is up to scale to predict a flow field of higher resolution. In order to provide dense per-pixel predictions, 'upconvolution' is performed on the coarse feature map to get it up to scale. 'Upconvolution' involves unpooling (bilinear upscaling of the feature map) followed by a convolution (refer Figure \ref{fig:exp1}(b)). Similar layers have been used previously \cite{dosovitskiy2015learning}. In this way, information from both coarser and finer feature maps is preserved. Upconvolution is performed at multiple scales which ultimately results in a two channel feature map which is 16 times the resolution of the last coarse feature map and 1/4 times the resolution of the image input. 
Our network differs slightly from FlowNet as we discard the loss layer of FlowNet and instead feed the final feature map to a fully connected layer with ReLu non linearity and dropout followed by separate regression layers for translation and rotation \cite{kendall2015posenet}.

\begin{figure*}[t!]
\begin{center}
\includegraphics[width=15.2cm, height=2.0cm] {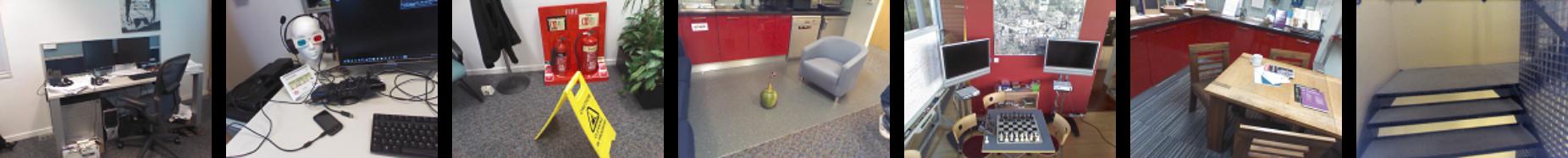} \\
\caption{\textbf{7 Scenes dataset}
Example images from left to right: Office, Heads, Fire, Pumpkin, Chess, Red Kitchen and Stairs.}
\label{fig:sevensceneds1}
\end{center}
\vspace{-0.3cm}
\end{figure*}
\section{Training}

\subsection{Loss function and Optimization Scheme}
Our network takes in two monocular images $I$, $I^*$ and outputs a pose vector $\textbf{p}$ comprising of a relative ($I^*$ with respect to $I$) translation $\textbf{x}$ and rotation $\textbf{q}$ in quaternion form. 
\begin{equation}
\mathbf{p=[x,q]}
\end{equation}

To regress relative pose, we consider the following objective loss function similar to \cite{kendall2015posenet}.
\begin{equation}
loss(I,I^*)= \left \| \hat{x}-x \right \|_{2} +\beta  \left \| \hat{q}-\frac{q}{\left \| q \right \|} \right \|_{2}
\end{equation}
$\beta$ is chosen so as to keep the expected value of translation and rotation errors to be equal. We found $\beta$ as $500,000$ to be optimal for training. The motive behind deploying the loss function is that both the translation and rotation regressors are loosely coupled and therefore, are not being denied the information to factor position from orientation and vice versa.

\subsection{Data-Preparation and Implementation Details}
We use the Train Split of 7 Scenes Dataset to train our networks. Ground truth is present for each frame in the form of $4\times4$ homogeneous matrix. For an image frame in a sequence, we take only 10 temporally close frames for computing the ground truth transformation, this is done to ensure that there is partial scene overlap in the two images. Let the absolute pose in homogeneous coordinates of $I$ and $I^*$ be $_{c}^{O}\textrm{T}$ and $_{c^*}^{O}\textrm{T}$ (with respect to some world origin $O$) respectively, then the Transformation of $I^*$ with respect to $I$ is given by:
\begin{equation}
_{c^*}^{c}\textrm{\textbf{T}} = _{O}^{c}\textrm{\textbf{T}} _{c^*}^{O}\textrm{\textbf{T}} 
\end{equation}
We obtain approximately 500,000 such training image pairs.
For training on FlowNet architecture, we resize the images to $512 \times 384$ and pass it for training. We use FlowNet's mean subtraction layer to normalize the image data.
We use Caffe library \cite{jia2014caffe} to train our networks. The machine has a core i7 processor with 64 GB of RAM. We used a single Titan X GPU to perform training and testing. It took an hour to complete 1000 iterations with a batch size of 40. 
We perform transfer learning \cite{yosinski2014transferable}, \cite{larochelle2009exploring} with the official FlowNet model weights released by the authors. This is done in order to get a better network initialization and faster network convergence. 
We use Adam optimization scheme instead of stochastic gradient descent for minimizing the loss function as it showed faster convergence for training during experiments. We train with base learning rate as $10^{-4}$ reduced by $50\%$ every $30,000$ iterations. We take momentum, momentum2 parameters of Adam solver to be $0.9$ and $0.99$ respectively. We use the network weights obtained after $75,000$ iterations of training for all our experiments.

\subsection{Dataset}
We train our network on the RGB-D '7 Scenes Dataset' \cite{glocker2013real}. It comprises of seven scenes of varying spatial extent and clutter as shown in figure \ref{fig:sevensceneds1}. The Dataset is challenging due to the presence of image artifacts such as motion blur and reflections. Also, presence of texture-less flat surfaces, sensor noise and varying lighting conditions compound the challenge. We chose this dataset as it comprises of multiple trajectories with a variety of rotational and translational transformations between frames. This would enable us to learn a rich variety of poses with challenging image pairs.  
\section{Control Law}
The network is trained to compute relative error in pose $_{c^*}^{c}\textrm{T}$ given two images $I$ and $I^*$. 
We consider an object centric coordinate system with a frame $\mathcal{F}_{o^*}$ attached to an object. $\mathcal{F}_{c}$,$\mathcal{F}_{c*}$ denote the current and desired camera frames. 
In our PBVS scheme, $\textbf{s}=(^{c^*}\textbf{t}_c, \theta \textbf{u})$
Consequently, $\textbf{s*}=\textbf{0}$ and $\textbf{e}=\textbf{s}$. This formulation results in a decoupling of rotational and translational motions and a simple control law as follows:

\begin{eqnarray}
\textbf{v}_c = & -{\lambda} { ^{c^*}}\textbf{R}^T_c  { ^{c^*}}\textbf{t}_c  \nonumber \\
\textbf{$\omega$}_c = & - {\lambda} \theta \textbf{u}.
\end{eqnarray}
Where, ${ ^{c^*}}\textbf{R}^T_c $ and ${ ^{c^*}}\textbf{t}_c$ are the relative rotation and translation of camera's desired pose with  respect to camera's initial pose in frame  $\mathcal{F}_{c}$ . ${ ^{c}}\textbf{R}^T_{c^*} $ and ${ ^{c}}\textbf{t}_{c^*}$ are predicted by our network, given $I$ and $I^*$. $\lambda$ is the step size for rotational and translational velocities. 

\section{Experiments and Results}
\label{sec:results}

\begin{figure*}[t!]
\begin{center}
 \begin{tabular}{|c|c|c|c|c|}
\hline
Initial pose & Desired pose & Resultant pose & Initial error image & Resultant error image\\
\hline
\vspace{0cm}\includegraphics[width=2.8cm, height=2.0cm] {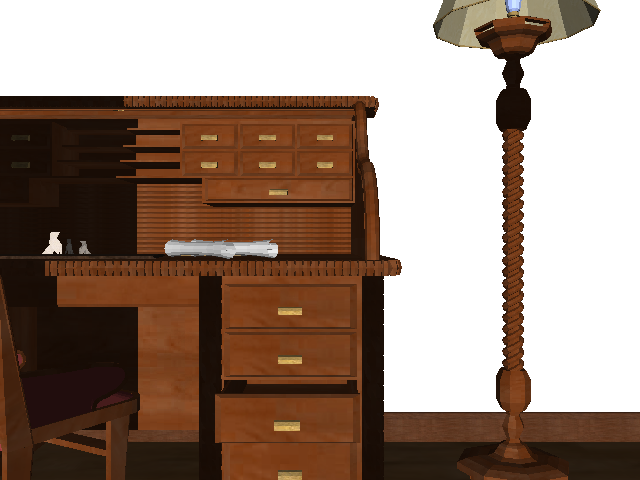} &
\vspace{0cm}\includegraphics[width=2.8cm, height=2.0cm] {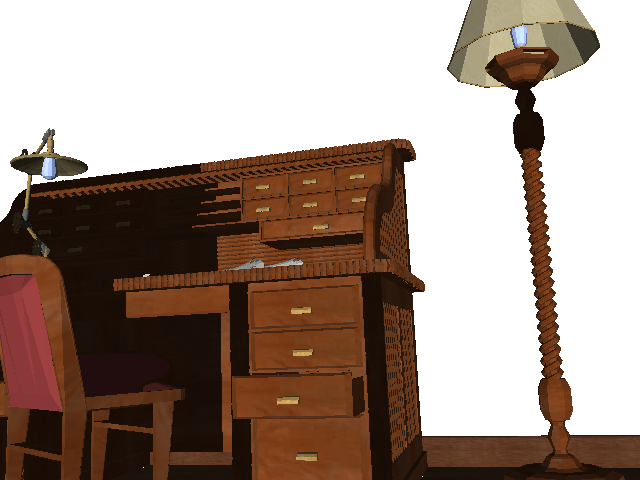} &
\vspace{0cm}\includegraphics[width=2.8cm, height=2.0cm] {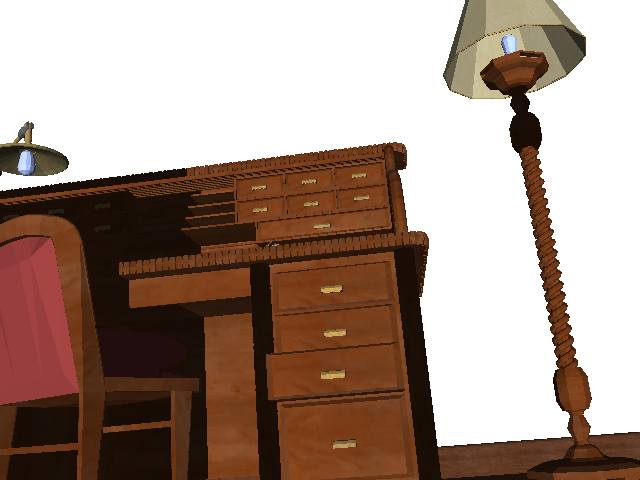} &
\vspace{0cm}\includegraphics[width=2.8cm, height=2.0cm] {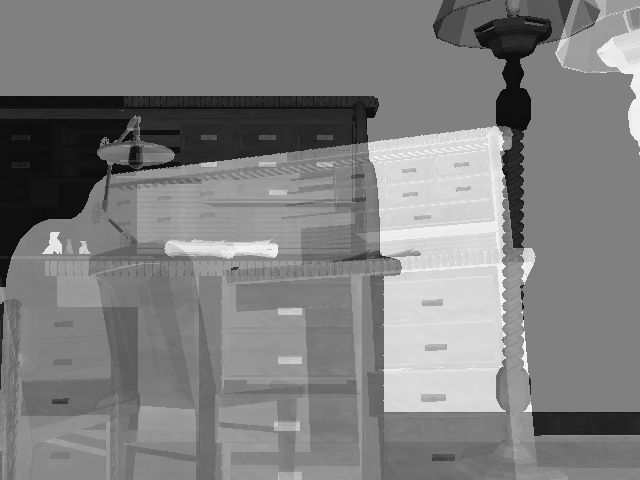} &
\vspace{0cm}\includegraphics[width=2.8cm, height=2.0cm] {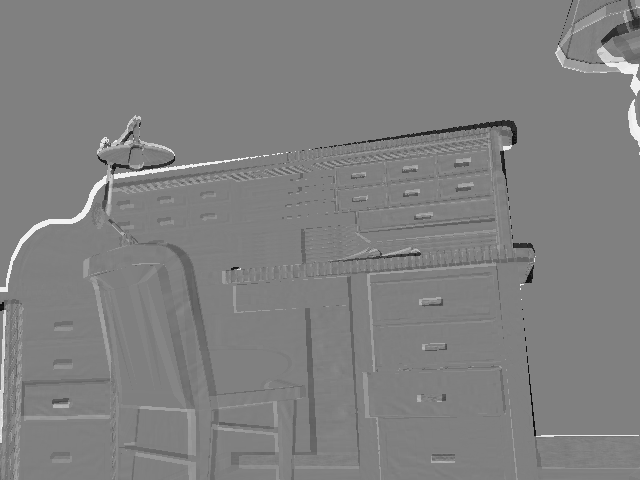} \\
\hline 
\vspace{0cm}\includegraphics[width=2.8cm, height=2.0cm] {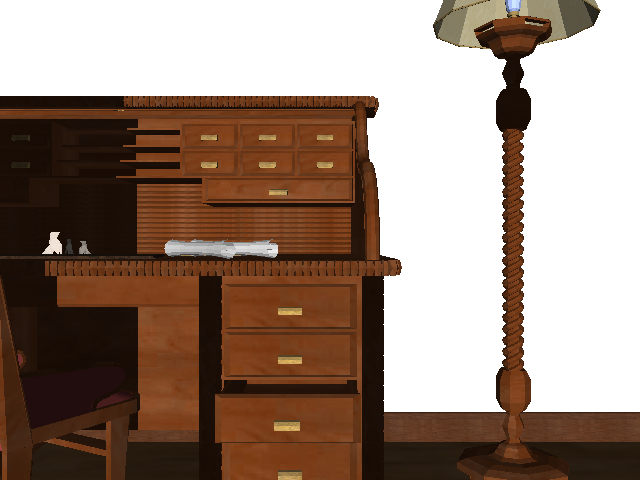} &
\vspace{0cm}\includegraphics[width=2.8cm, height=2.0cm] {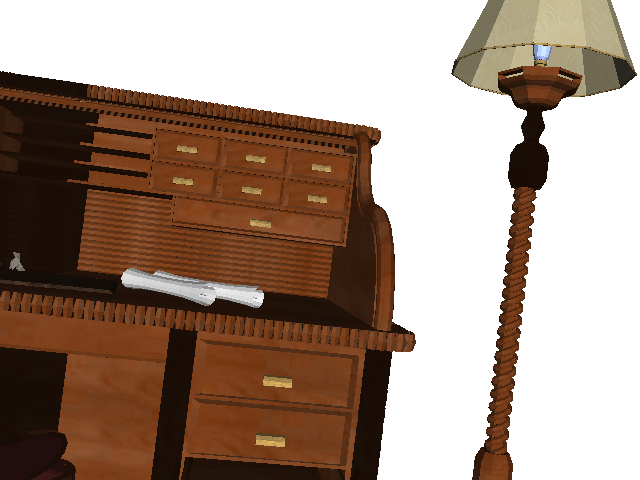} &
\vspace{0cm}\includegraphics[width=2.8cm, height=2.0cm] {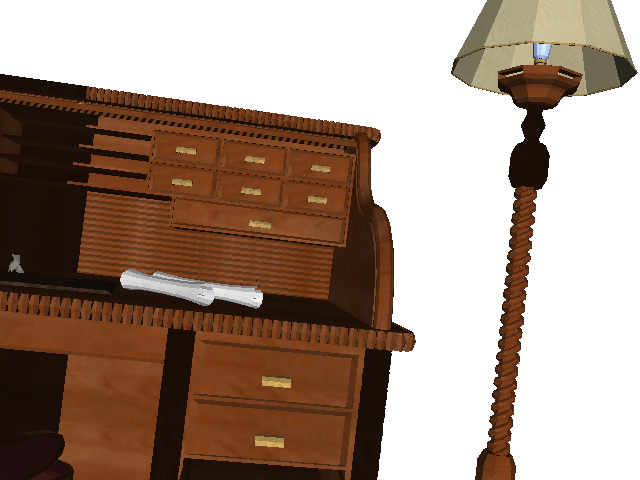} &
\vspace{0cm}\includegraphics[width=2.8cm, height=2.0cm] {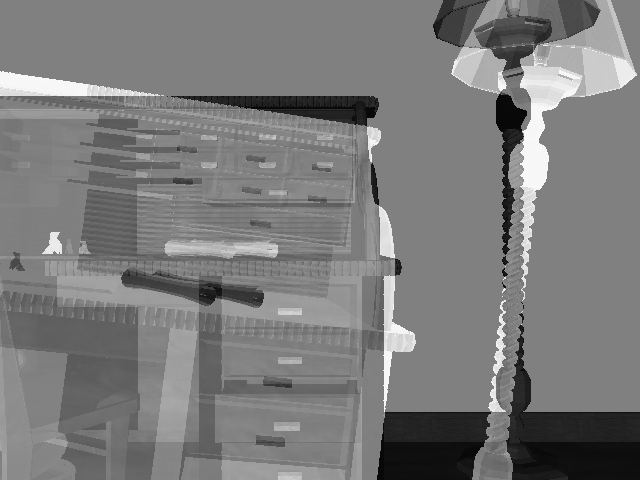} &
\vspace{0cm}\includegraphics[width=2.8cm, height=2.0cm] {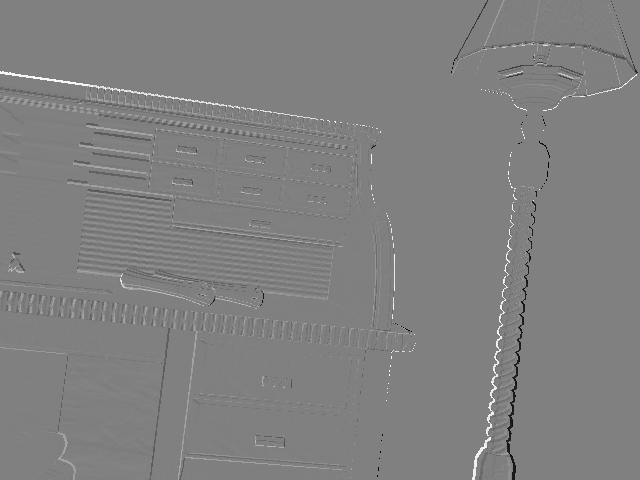} \\
\hline 
\rule{0pt}{8ex} 
\vspace{0cm}\includegraphics[width=2.8cm, height=2.0cm] {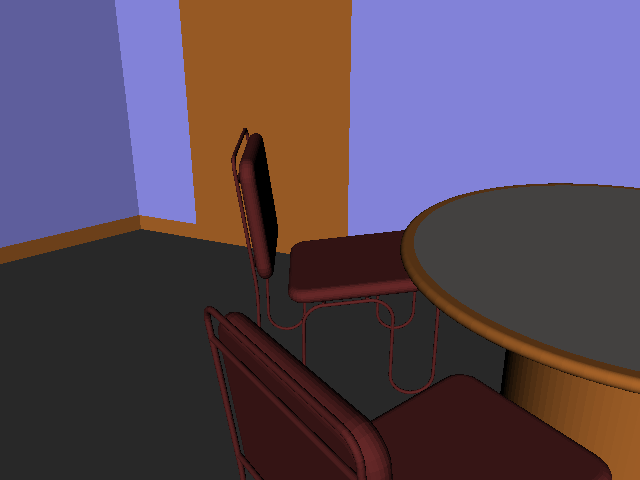} &
\vspace{0cm}\includegraphics[width=2.8cm, height=2.0cm] {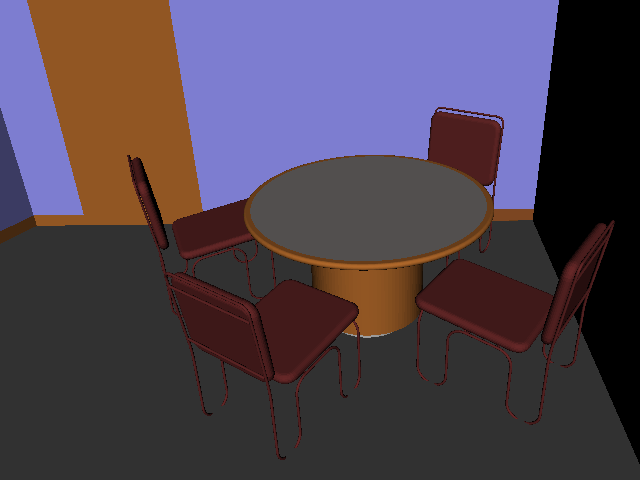} &
\vspace{0cm}\includegraphics[width=2.8cm, height=2.0cm] {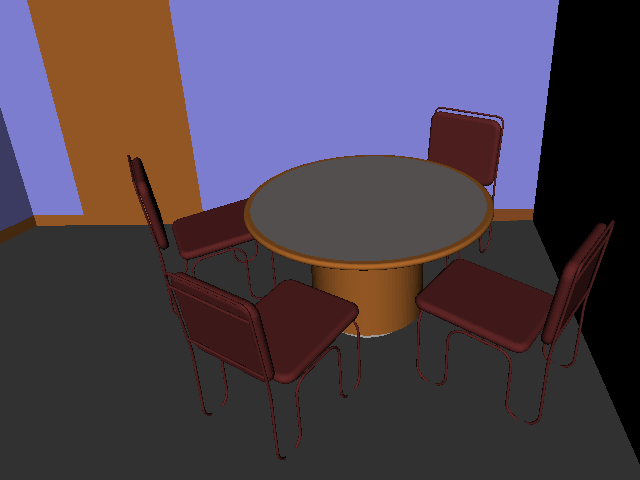} &
\vspace{0cm}\includegraphics[width=2.8cm, height=2.0cm] {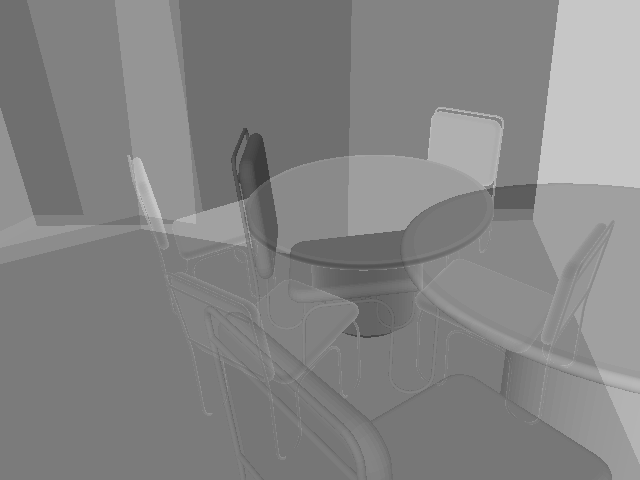} &
\vspace{0cm}\includegraphics[width=2.8cm, height=2.0cm] {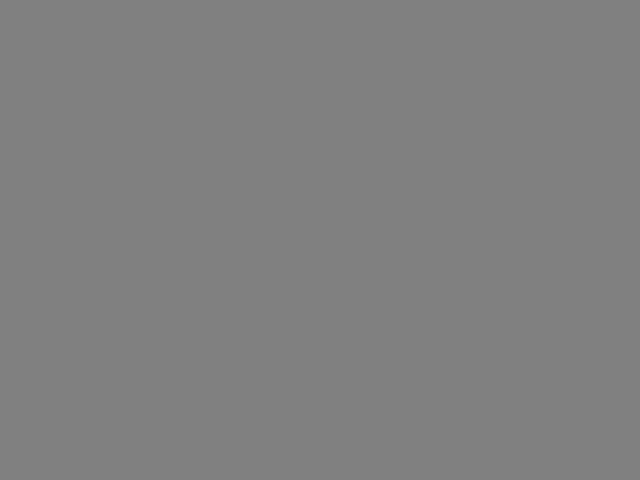} \\
\hline 
\rule{0pt}{8ex} 
\vspace{0cm}\includegraphics[width=2.8cm, height=2.0cm] {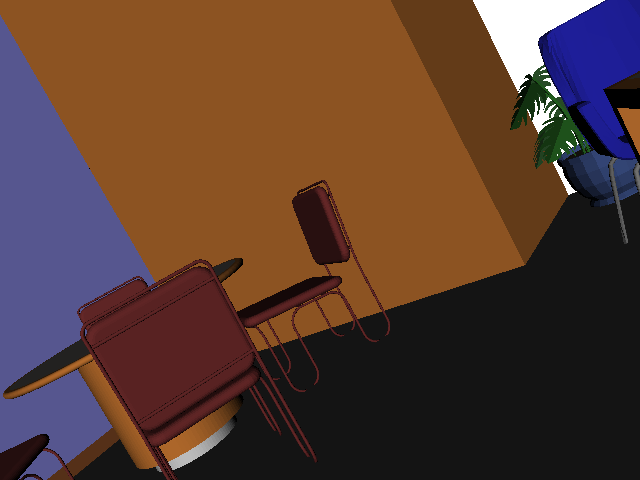} &
\vspace{0cm}\includegraphics[width=2.8cm, height=2.0cm] {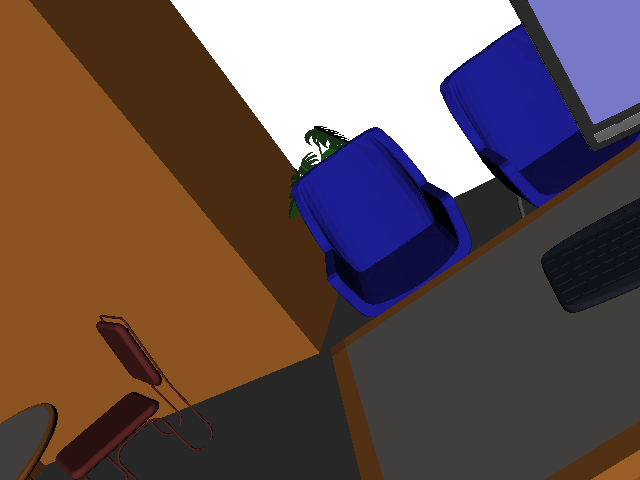} &
\vspace{0cm}\includegraphics[width=2.8cm, height=2.0cm] {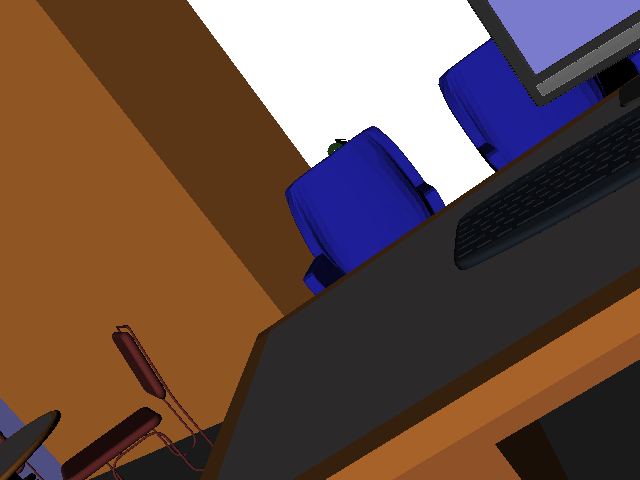} &
\vspace{0cm}\includegraphics[width=2.8cm, height=2.0cm] {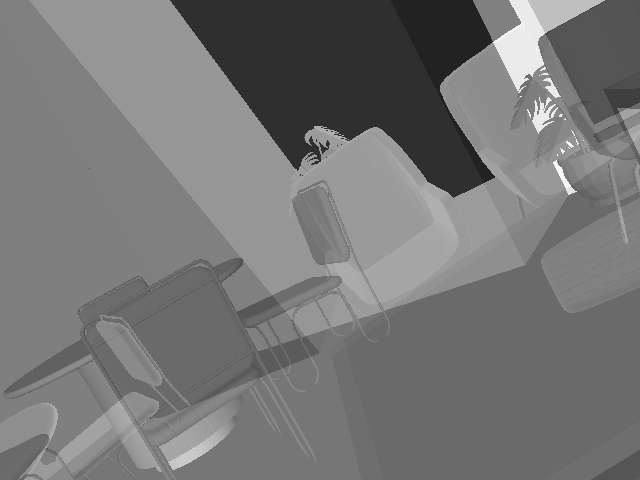} &
\vspace{0cm}\includegraphics[width=2.8cm, height=2.0cm] {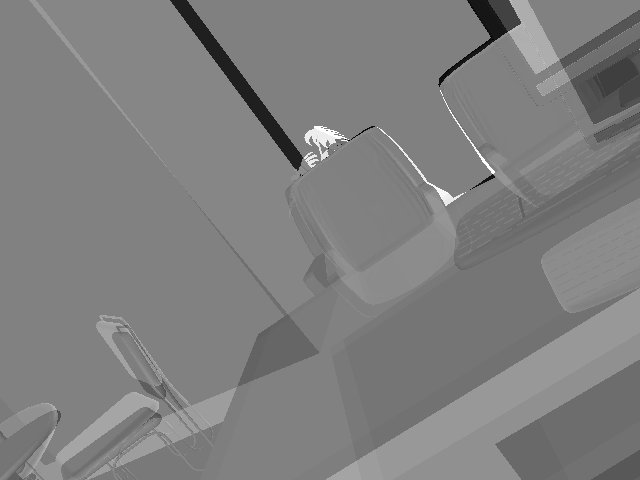} \\
\hline 
\rule{0pt}{8ex} 
\vspace{0cm}\includegraphics[width=2.8cm, height=2.0cm] {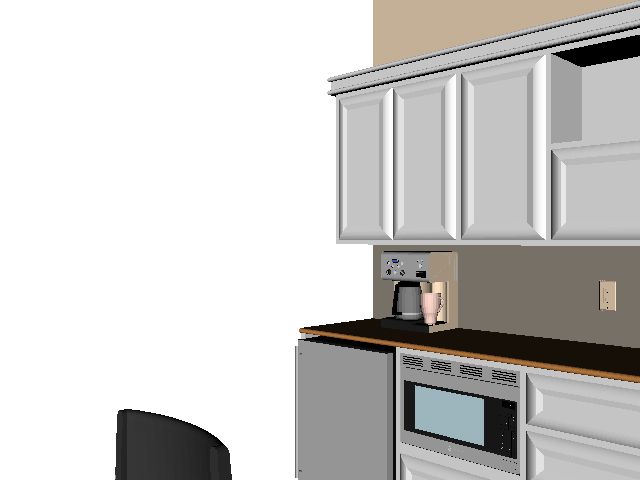} &
\vspace{0cm}\includegraphics[width=2.8cm, height=2.0cm] {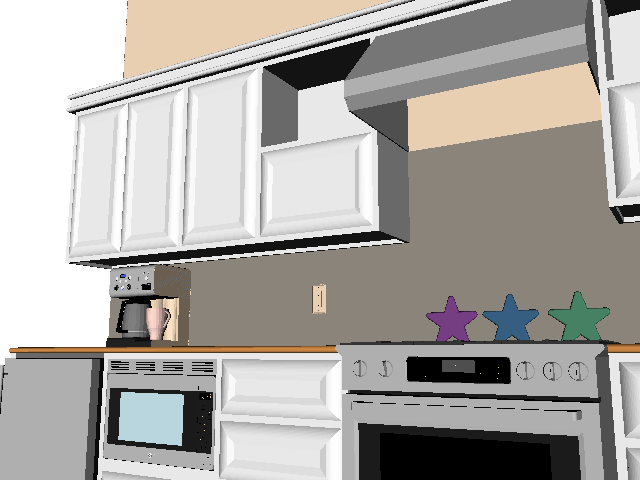} &
\vspace{0cm}\includegraphics[width=2.8cm, height=2.0cm] {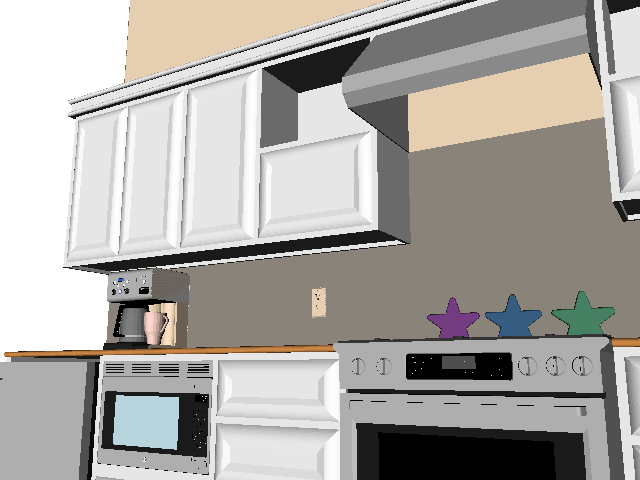} &
\vspace{0cm}\includegraphics[width=2.8cm, height=2.0cm] {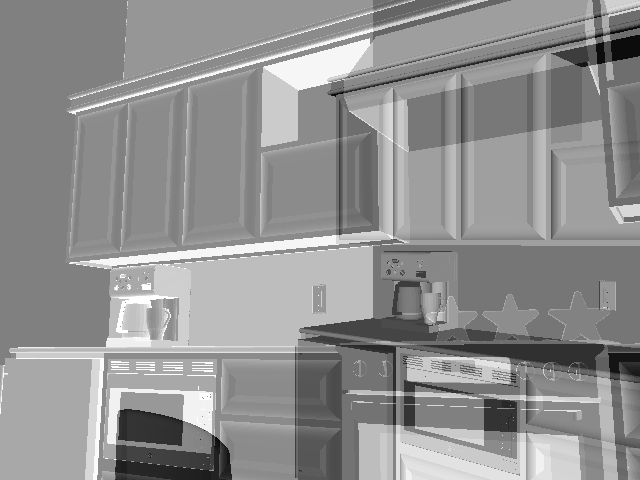} &
\vspace{0cm}\includegraphics[width=2.8cm, height=2.0cm] {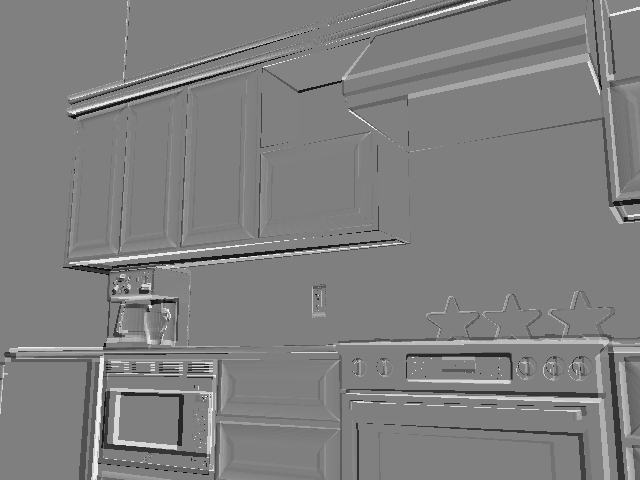} \\
\hline 
\rule{0pt}{8ex} 
\vspace{0cm}\includegraphics[width=2.8cm, height=2.0cm] {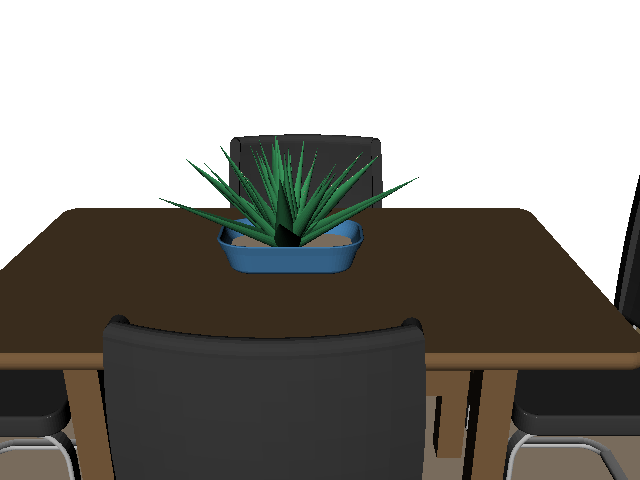} &
\vspace{0cm}\includegraphics[width=2.8cm, height=2.0cm] {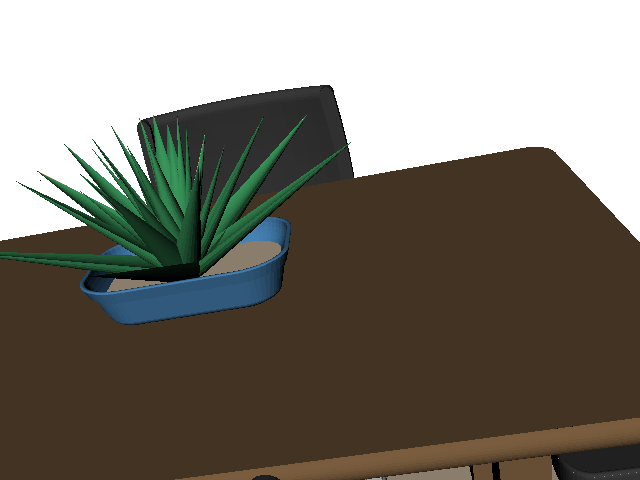} &
\vspace{0cm}\includegraphics[width=2.8cm, height=2.0cm] {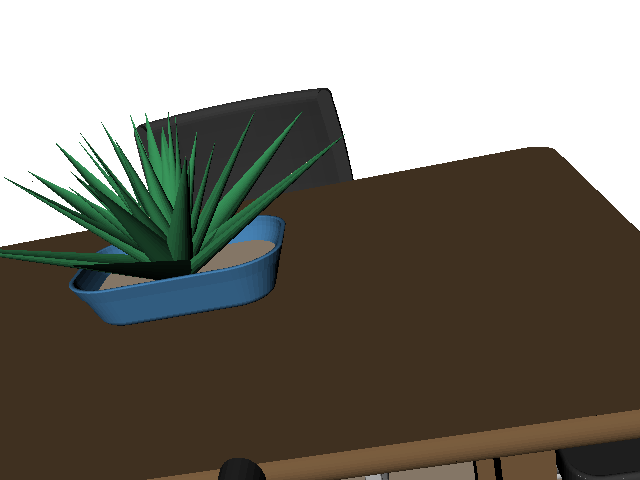} &
\vspace{0cm}\includegraphics[width=2.8cm, height=2.0cm] {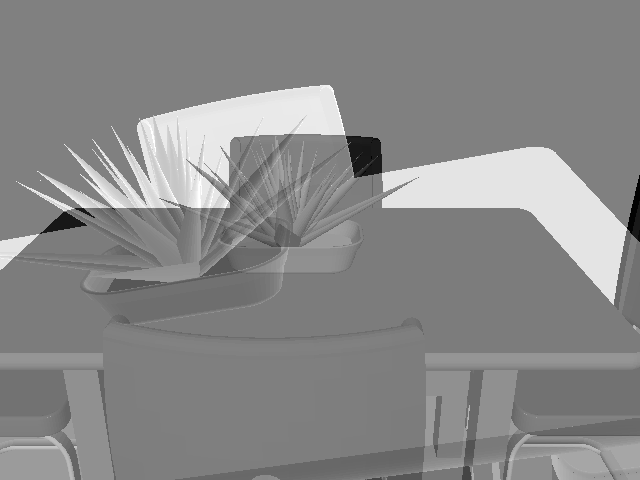} &
\vspace{0cm}\includegraphics[width=2.8cm, height=2.0cm] {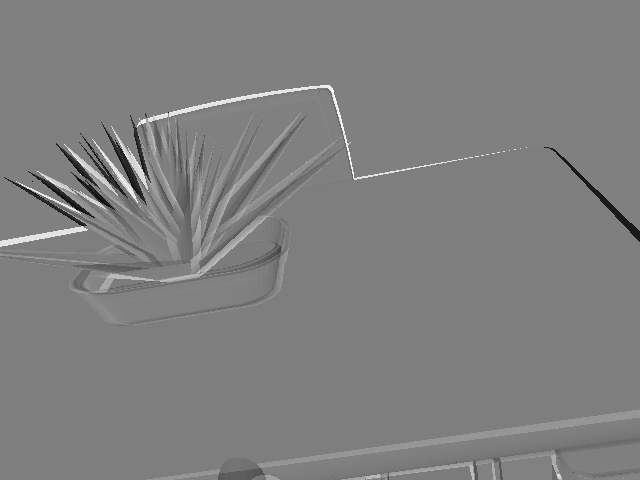} \\
 \hline 
 \rule{0pt}{8ex} 
 \vspace{0cm}\includegraphics[width=2.8cm, height=2.0cm] {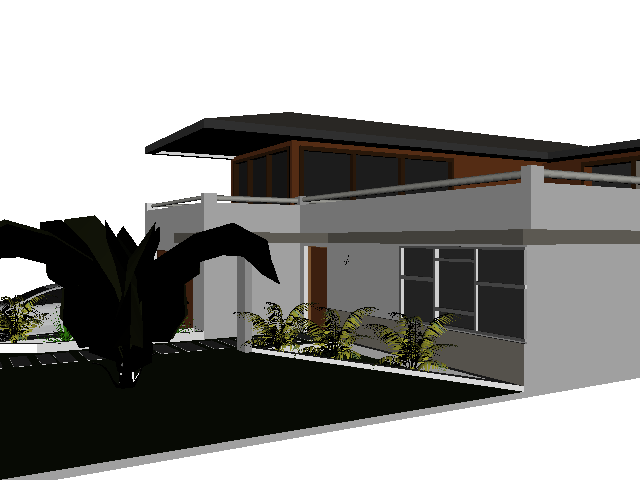} &
 \vspace{0cm}\includegraphics[width=2.8cm, height=2.0cm] {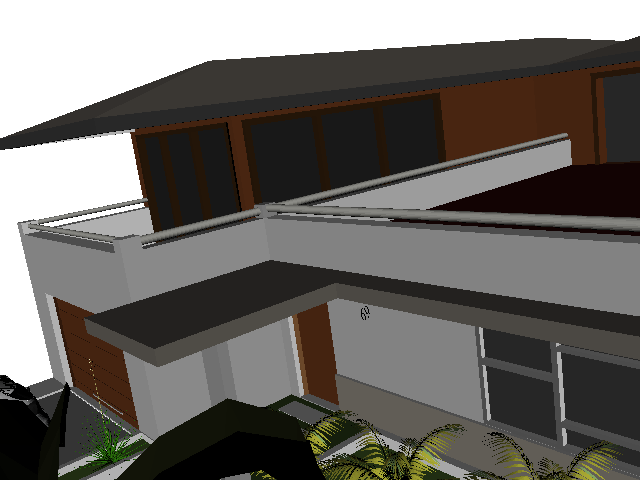} &
 \vspace{0cm}\includegraphics[width=2.8cm, height=2.0cm] {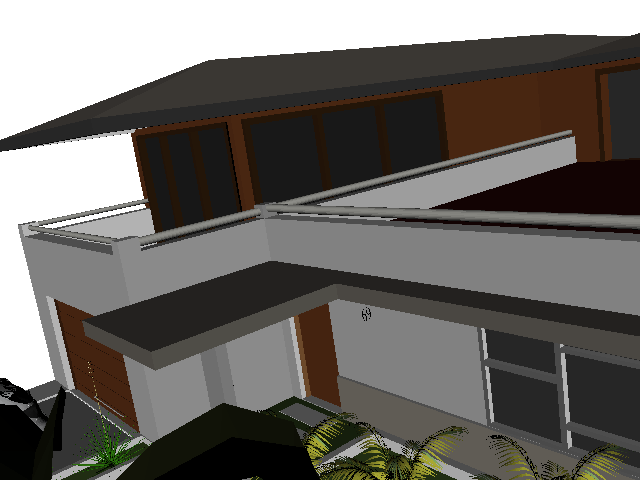} &
 \vspace{0cm}\includegraphics[width=2.8cm, height=2.0cm] {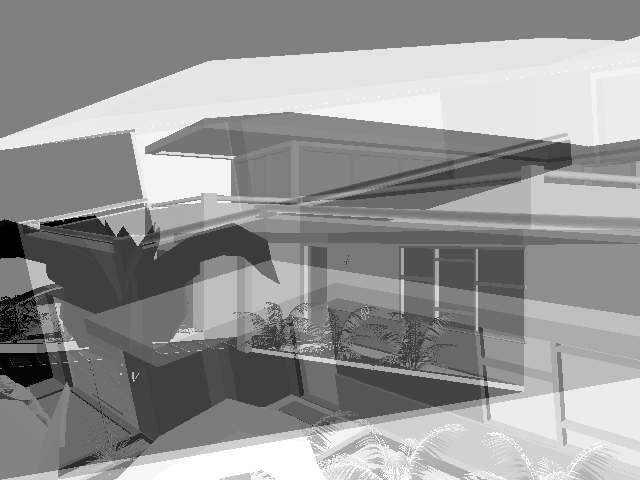} &
 \vspace{0cm}\includegraphics[width=2.8cm, height=2.0cm] {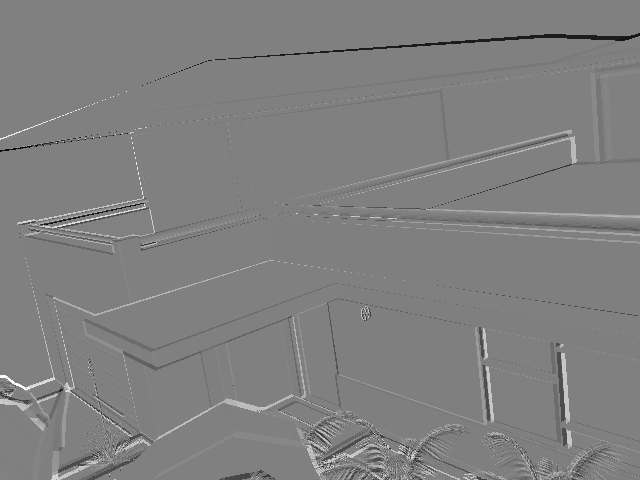} \\
\hline 
\rule{0pt}{8ex} 
\vspace{0cm}\includegraphics[width=2.8cm, height=2.0cm] {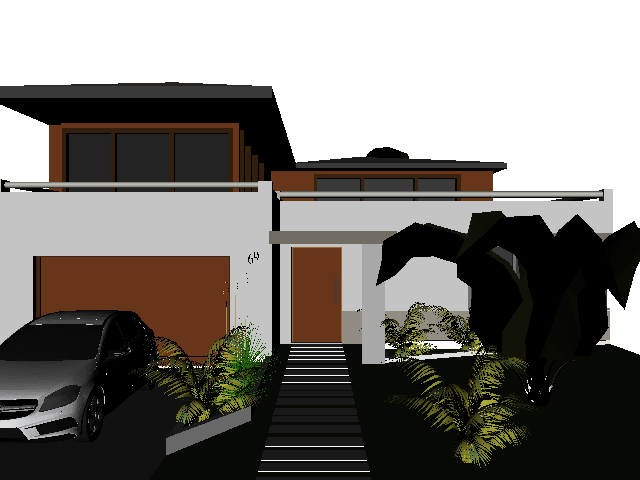} &
\vspace{0cm}\includegraphics[width=2.8cm, height=2.0cm] {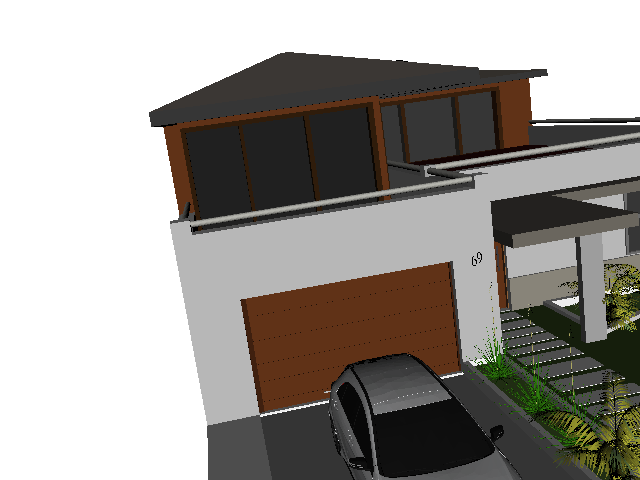} &
\vspace{0cm}\includegraphics[width=2.8cm, height=2.0cm] {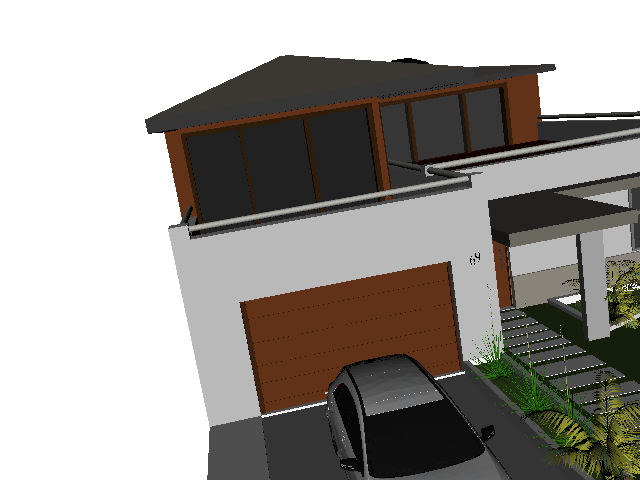} &
\vspace{0cm}\includegraphics[width=2.8cm, height=2.0cm] {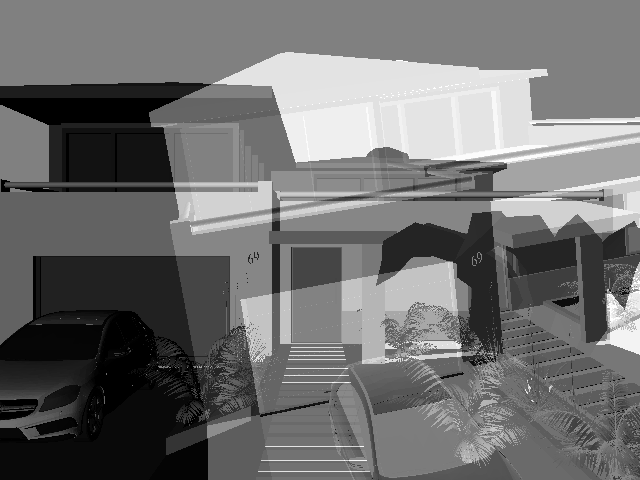} &
\vspace{0cm}\includegraphics[width=2.8cm, height=2.0cm] {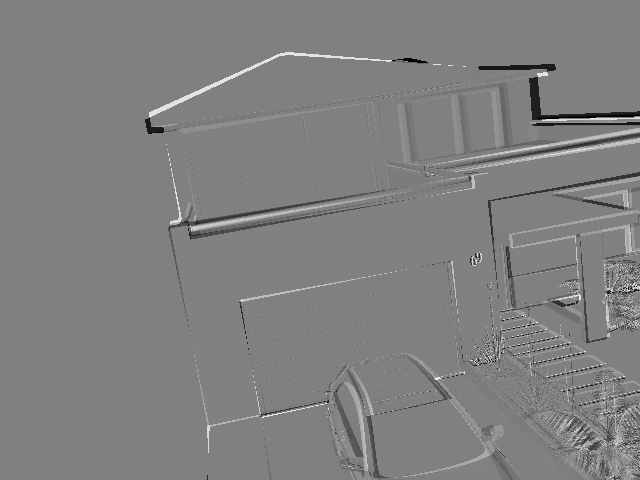} \\
\hline 
\rule{0pt}{8ex} 
\vspace{0cm}\includegraphics[width=2.8cm, height=2.0cm] {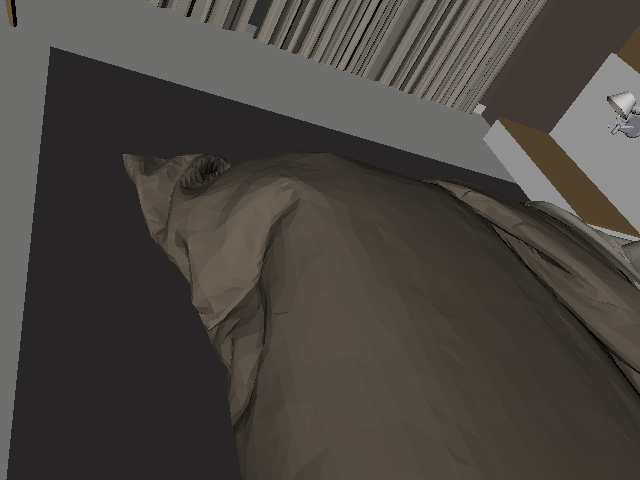} &
\vspace{0cm}\includegraphics[width=2.8cm, height=2.0cm] {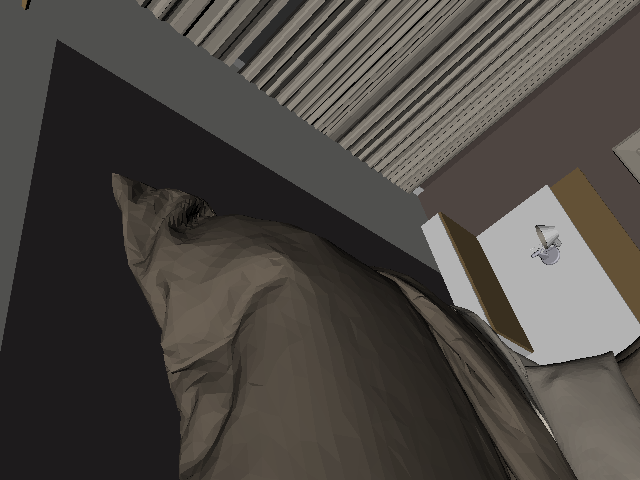} &
\vspace{0cm}\includegraphics[width=2.8cm, height=2.0cm] {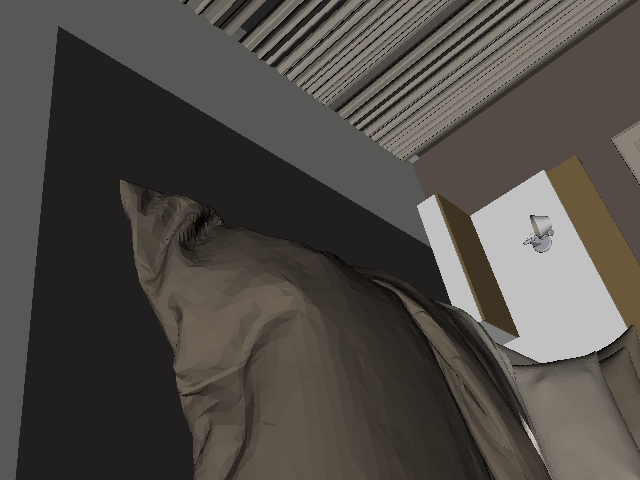} &
\vspace{0cm}\includegraphics[width=2.8cm, height=2.0cm] {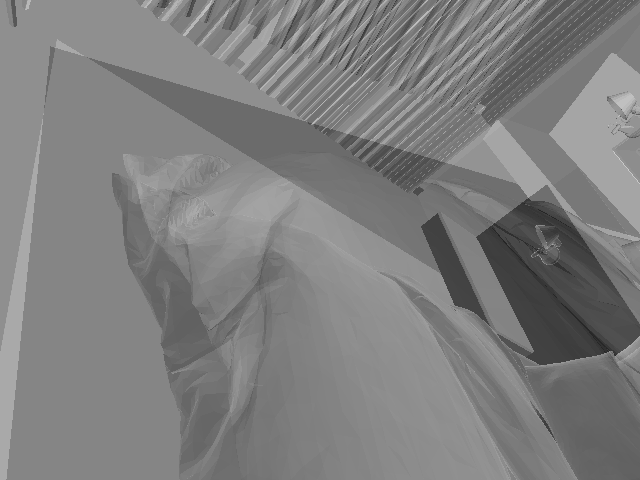} &
\vspace{0cm}\includegraphics[width=2.8cm, height=2.0cm] {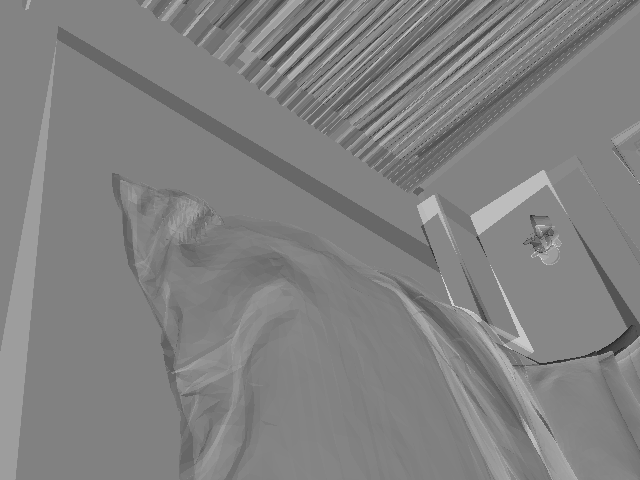} \\
\hline 
\rule{0pt}{8ex} 
\vspace{0cm}\includegraphics[width=2.8cm, height=2.0cm] {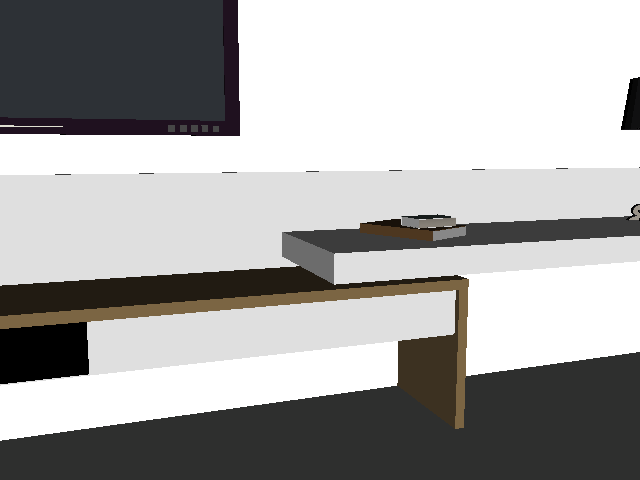} &
\vspace{0cm}\includegraphics[width=2.8cm, height=2.0cm] {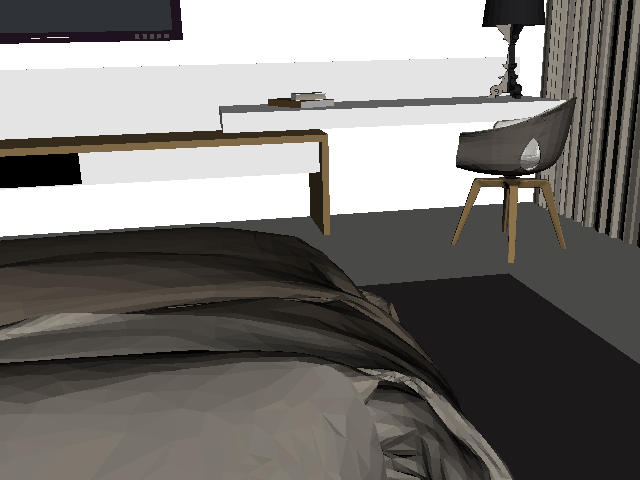} &
\vspace{0cm}\includegraphics[width=2.8cm, height=2.0cm] {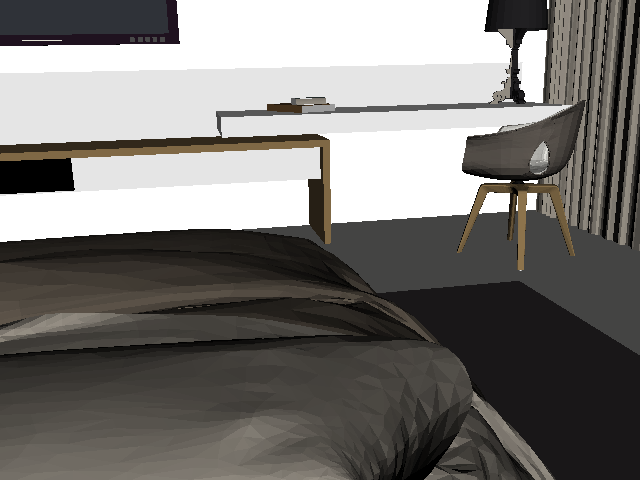} &
\vspace{0cm}\includegraphics[width=2.8cm, height=2.0cm] {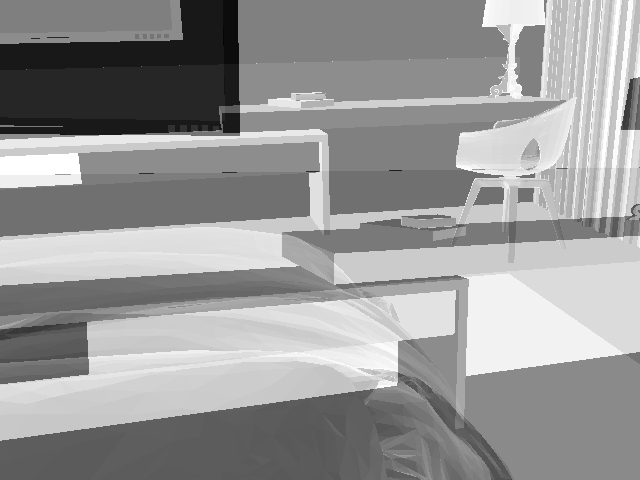} &
\vspace{0cm}\includegraphics[width=2.8cm, height=2.0cm] {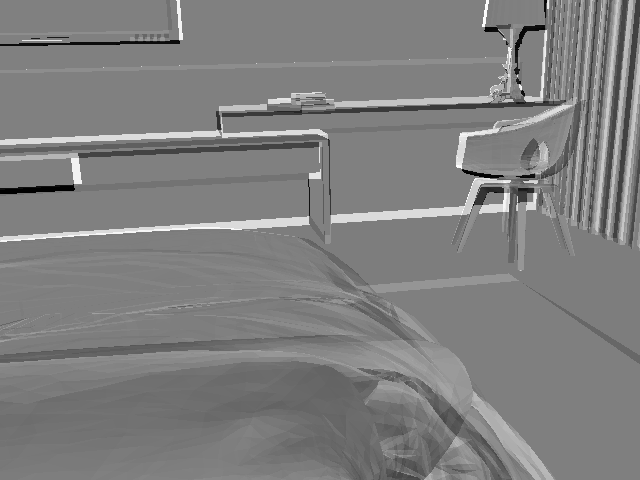} \\
\hline 
\end{tabular}
\end{center}

\caption{\textbf{Qualitative results.} (a) Initial pose captured by the robot with a random camera pose for provided $3$D object from the dataset. (b) Desired pose. (c) Resultant positioning of camera achieved by CNN based visual servoing. (d) Resultant error image. Note the similarity in the resultant pose achieved by the proposed approach compared to the desired pose provided, over a wide range of desired poses. }
\label{fig:qualitative}

\end{figure*}
 We evaluate our approach on non-planar scenes. Since there is no publicly available dataset that allows us to render an entire scene from a given viewpoint, we introduce a new synthetic dataset VSSD consisting of $5$ detailed CAD models of various scenes. We use the OpenRAVE simulation framework \cite{openrave} for rendering scenes since it  allows us to quantitatively measure the performance of our approach as the desired camera pose is known in the world frame. Thirdly, we qualitatively evaluate the performance of our approach on our dataset for various initial and desired poses. Finally, we execute our approach in a real environment using a quadrotor. Note that for all the experiments we do not assume any knowledge of camera parameters or depth information of the scene. Another fact worth noting is that the images used in evaluation were not encountered during training of the CNN Model. For simulation experiments we consider free-flying camera model. All the experiments reported here are performed on a system with Intel i7 processor and 64 GB RAM and a single 12 GB Titan X GPU. On this system our approach takes $20$s for initialization i.e. loading the network into GPU and henceforth every iteration takes $20$ms of which, majority of the time is consumed in forward pass of the network.  

\subsection{Visual Servoing Scene Dataset}
There are several publicly available datasets for tracking and localization \cite{geiger2012we}, \cite{glocker2013real}. However, for visual servoing it is difficult to release such a dataset, as it requires image based measurements of the environment where viewpoint changes dynamically. We address this issue by using synthetic $3$D models. In the recent past, $3$D models have been used by the computer graphics and vision communities to produce large amounts of synthetic data which enable better generalisation for deep learning models \cite{shilane2004princeton}. However, these datasets are limited to shapes and objects. Recently Handa et al. \cite{handa2015scenenet} released a synthetic dataset for scenes. However, the scenes provided are purely depth-based, which makes it unsuitable for visual servoing purposes. $3$D positioning is performed for physical objects which limits the scope of reproducing the results for benchmarking purposes.\\
\indent For this work, we have generated Visual Servoing Scene Dataset (VSSD) by rendering $5$ scenes using textured synthetic $3$D models publicly available from Google $3$D warehouse \cite{google3dwarehouse}. We have ensured to diversify scenes based on the following criterion:
\begin{itemize}
    \item We have selected models that represent indoor, outdoor and object categories.
    \item The scenes are sufficiently large to capture large camera transformations.
    \item These scenes have non-homogeneous lightning conditions.
    \item Viewpoints in the scenes vary in texture.
\end{itemize}
\indent The main motivation behind the effort is to provide a wide range of test cases for systematically evaluating visual servoing approaches on a common benchmark. All the scenes (CAD models) used in the dataset are publicly available and could be download at our project page\footnote{\url{http://robotics.iiit.ac.in/urls/d173716a.htm}}.

\subsubsection{Simulation results for $3$D scene}
\begin{figure}[h!]
\begin{center}
\begin{tabular}{cc}
\framebox{\includegraphics[width=2.8cm, height=1.7cm] {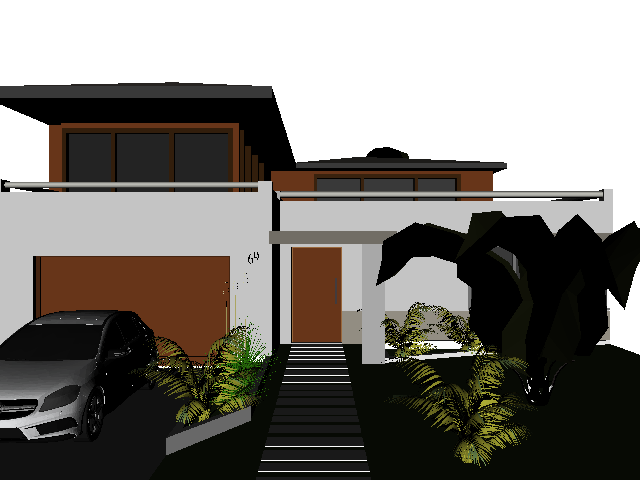}} &
\framebox{\includegraphics[width=2.8cm, height=1.7cm] {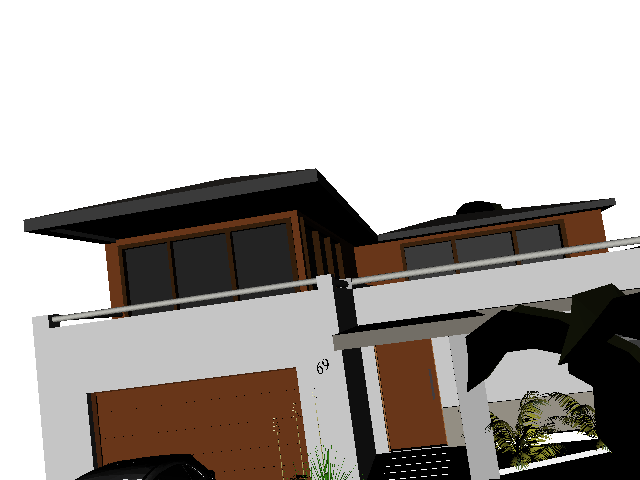}} \\
(a) & (b) \\
\includegraphics[width=3cm, height=1.8cm] {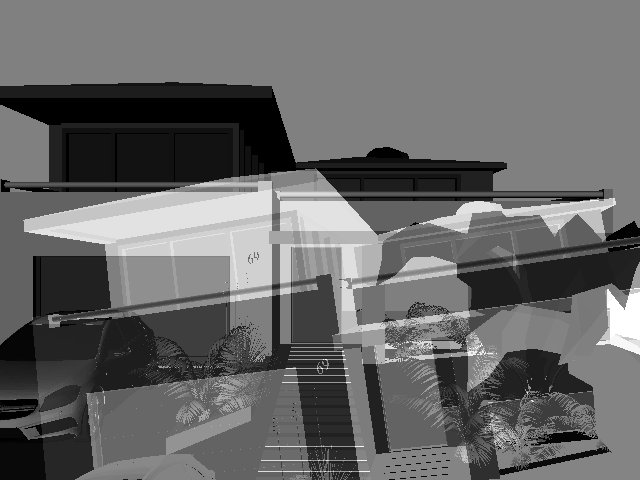} &
\includegraphics[width=3cm, height=1.8cm] {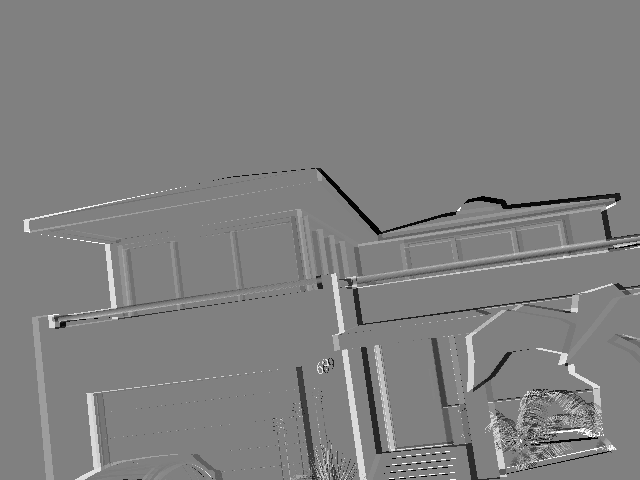} \\
(c) & (d) \\
\includegraphics[width=3.8cm, height=2.4cm] {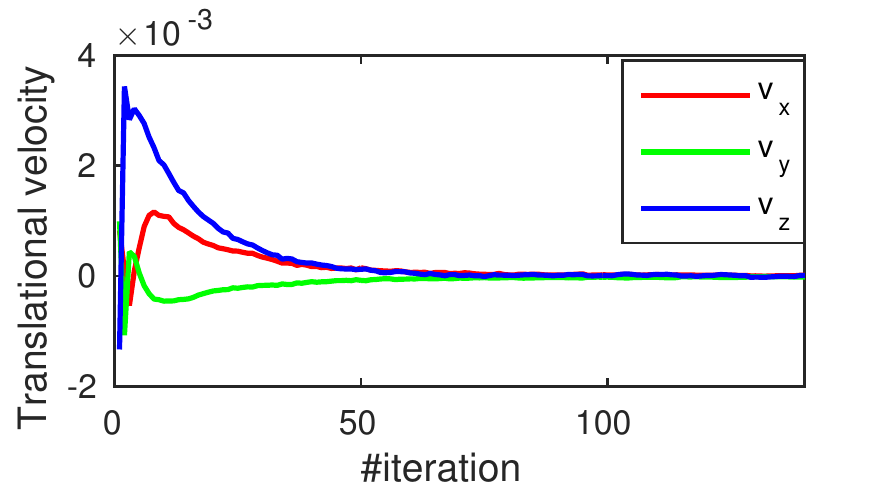} &
\includegraphics[width=3.8cm, height=2.4cm] {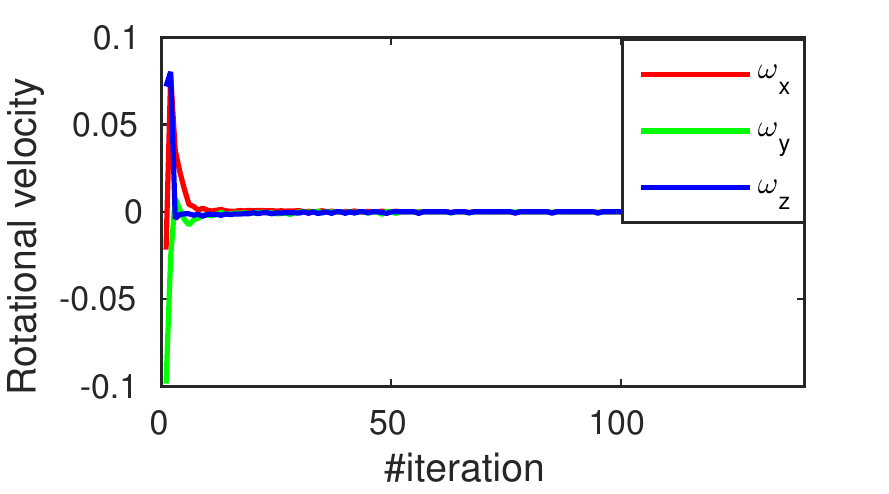} \\
(e) & (f) \\
\includegraphics[width=3.8cm, height=2.4cm] {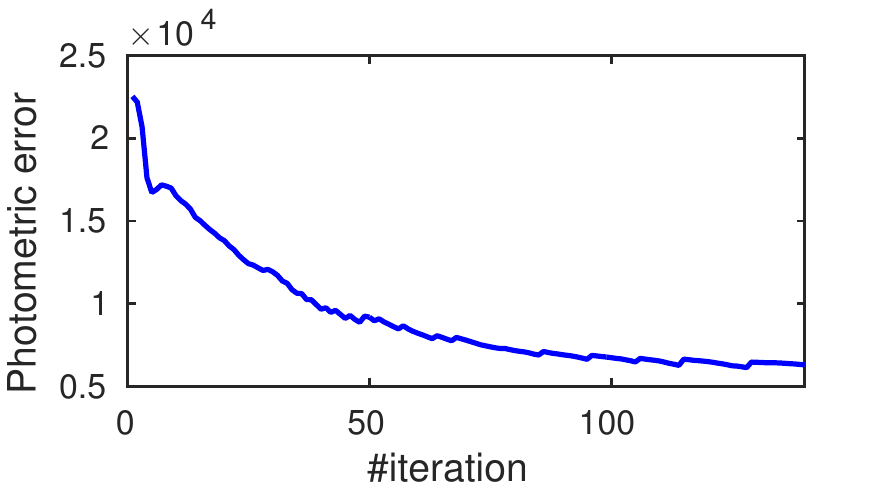} &
\includegraphics[width=3.8cm, height=2.4cm] {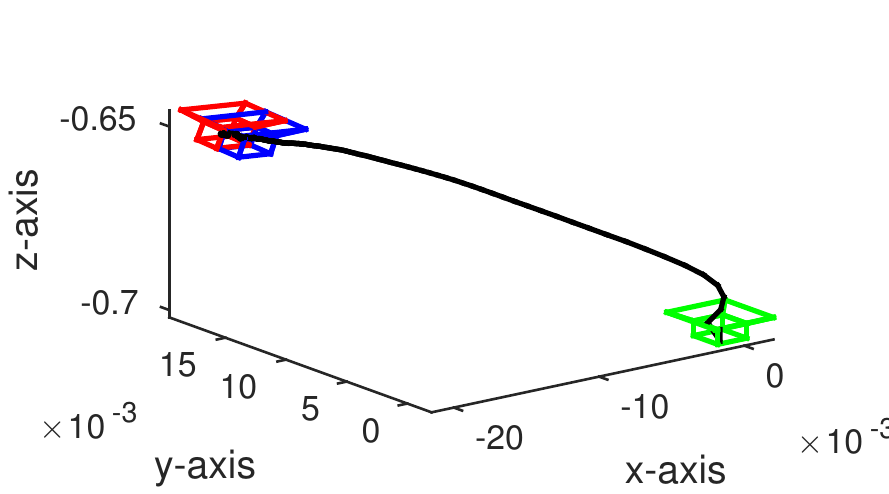} \\
(g) & (h)
\end{tabular}
\end{center}
\caption{\textbf{3$D$ positioning task.} (a) Initial pose for a non-planar scene from house scene. (b) Desired pose. (c) Error image for initial image. Notice the large displacement of camera and variation in viewpoints. (d) Error image for resultant pose using CNN. 
(e) Translational velocity in m/s. (f) Rotational velocity in rad/s. (g) Photometric feature error. (h) Camera trajectory. Our approach is able to attain the desired pose even when the displacement between initial and desired pose is large and lightning is non-homogeneous.}
\label{fig:exp2}
\vspace{-0.50em}
\end{figure}

In this experiment we aim to evaluate the control laws for our network architecture and to evaluate robustness in performing a positioning task.
The initial pose (refer figure \ref{fig:exp2}(a)) is selected from "House" model of VSSD dataset. The difference between desired and initial pose $\Delta r_{\text{desired}} = r_{\text{desired}} -r_{\text{init}} =[91.4 \text{mm, }   92.3\text{mm, }    36.7\text{mm, }    8\text{\degree },   10\text{\degree ,}   -5 \text{\degree } ]$. Although, the relative camera transformation is large, our approach is still able to converge to the desired pose with error in camera pose as $\Delta  r_{\text{desired}} - \Delta r_{\text{final}} =  [-5.1 \text{mm, }     2.8 \text{mm, }    0.5 \text{mm, }    -0.28 \text{\degree ,}  -0.42 \text{\degree }  -0.42  \text{\degree ,}]$, which is around $4 \%$ in both translation as well as rotation. It could be seen from figure \ref{fig:exp2}(e-g) that both the error as well as the camera velocity decrease exponentially despite the fact that these are output by a CNN. The experiment demonstrates that visuomotor representations are indeed learnt by our system. Also, the camera trajectory as shown in figure \ref{fig:exp2}(h)is close to a straight line, which is desirable for visual servoing purposes.

\subsubsection{Qualitative results on servoing dataset}
The objective of this experiment is to show the efficacy of the proposed algorithm to servo to a diverse set of target instances across various environment and viewpoint variations. For every scene from the VSSD dataset, we evaluate our algorithm for two configurations of the initial and desired pose pair, with different transformations in $6$ DOF. The resultant error images from figure \ref{fig:qualitative} indicate that our CNN based approach is indeed able to attain the desired pose for large camera pose variations. Let us note that VSSD has non-homogeneous lighting conditions, hence the assumption of temporal luminance continuity made by previous featureless visual servoing approaches \cite{photometricvs}, \cite{gradientvs} does not apply to such scenes. Also, the scene "kitchen" has textureless surfaces, which would make feature extraction difficult. This experiment validates the robustness of the feature representations learnt by the network for diverse and challenging environments without prior knowledge of the scene or camera used. 

\subsubsection{Real experiment using a quadrotor}

\begin{figure}[h!]
\begin{center}
\begin{tabular}{cc}
\includegraphics[width=3.8cm, height=2.2cm] {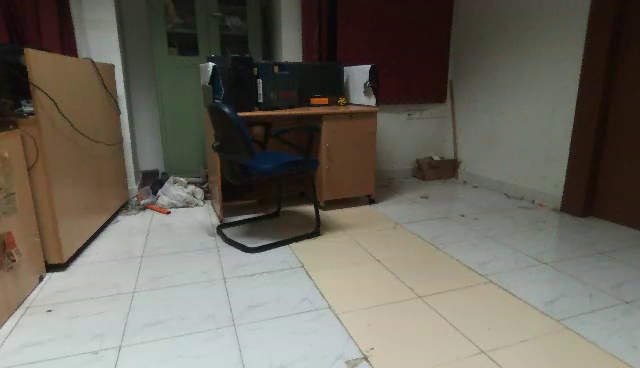} &
\includegraphics[width=3.8cm, height=2.2cm] {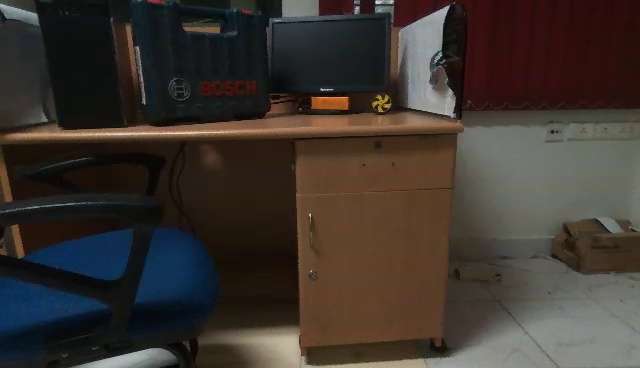} \\
(a) & (b) \\
\includegraphics[width=3.8cm, height=2.2cm] {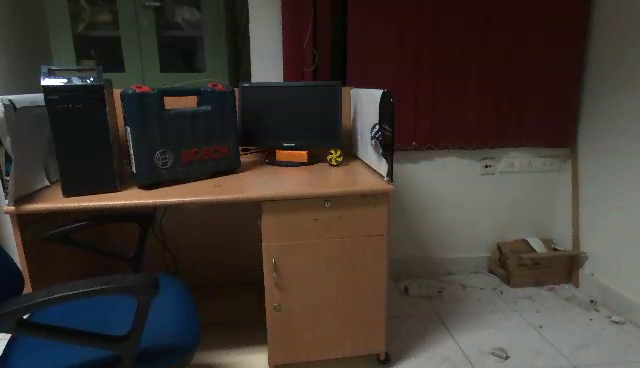} &
\includegraphics[width=3.8cm, height=2.2cm] {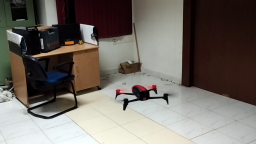} \\
(c) & (d) \\
\includegraphics[width=3.8cm, height=2.2cm] {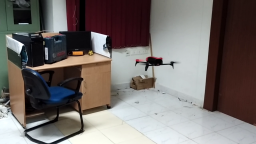} & 
\includegraphics[width=3.8cm, height=2.4cm] {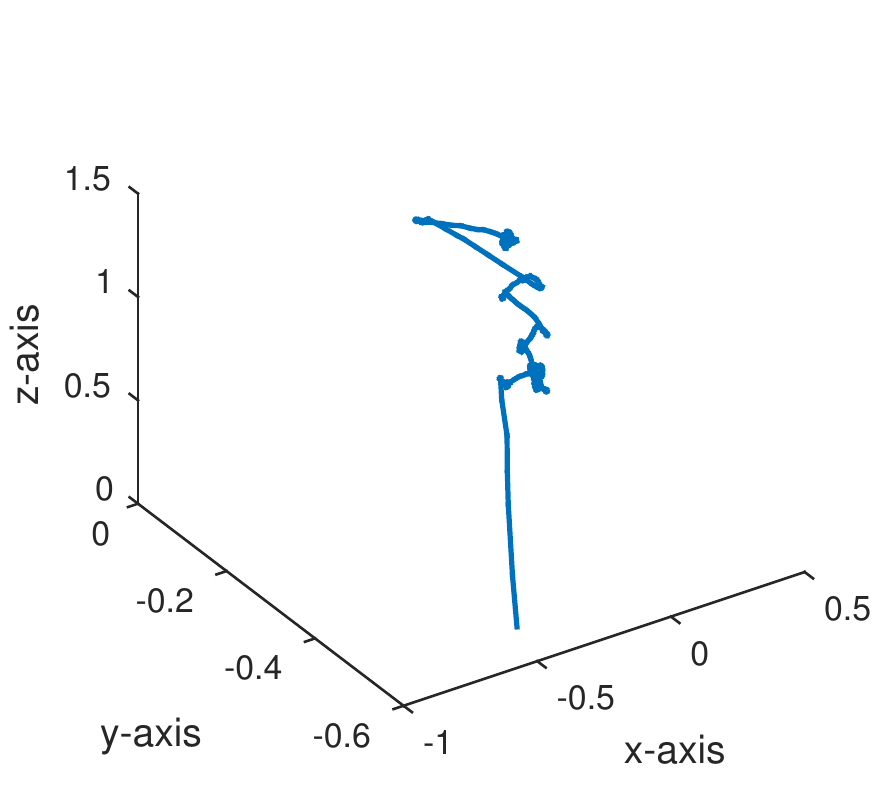}
\\
(e) & (f)
\end{tabular}
\end{center}
\caption{\textbf{Positioning task using quadrotor.} (a) Initial pose for a real scene. (b) Desired pose. (c) Resultant pose at the end of approach. Notice the large displacement of camera and variation in viewpoints. (d) Initial position of quadrotor in the image space. (e) Final position of quadrotor in image space. (f) Approximate quadrotor trajectory in 3D.}
\label{fig:quad}
\vspace{-0.50em}
\end{figure}

In this experiment, we evaluate our approach on real world scenarios using a Parrot Bebop 2 drone. Since quadrotors are under-actuated, only $4$ DOF tasks were selected for visual servoing. In real world, it is difficult to accurately predict the position of a drone. Hence we report the qualitative results and an approximate trajectory generated and reported by the drone by fusion of inertial measurement unit (IMU) , sonar sensor and optical flow sensor facing downward. Note that the images in the evaluation were not encountered during training of the CNN model. Again, the transformation between the initial and the desired pose is large. Precise convergence was not achieved since only $4$ DOF could be controlled. Figure \ref{fig:quad}(a,b) show the initial and desired pose given to the CNN for generating velocity commands. The local controller aimed to track the quadrotor velocity commands generated by the CNN based  high-level controller. The CNN forward pass processing was performed using a laptop computer with Core i7 CPU, Nvidia Quadro M2000M GPU and 16 GB RAM. It took $65$ms for one forward pass to complete on the machine. The drone was given 2 seconds to converge to the generated velocity before capture and forward pass of next image hence sending next velocity command. The image captured by the drone and corresponding control commands generated by the network were exchanged between the system and drone over WiFi channel.
\section{Conclusion}
In this work, we have introduced an end-to-end learning based framework for visual servoing tasks using CNN. The visuomotor representations learnt by the network generalises well across diverse environments. We have experimentally verified our approach on both synthetic as well as real world scenarios for robustness to non-homogeneous illumination and texture of scene. Unlike previous approaches, we do not need the knowledge of geometry of scene or camera parameters. Further, by learning the control representations we circumvent the requirement of any feature extraction or tracking step.
\bibliographystyle{IEEEtran}
\bibliography{egbib2}

\end{document}